\documentclass[10pt,twocolumn,letterpaper]{article}

\usepackage{iccv}              
\usepackage[accsupp]{axessibility}  
\usepackage{makecell}
\usepackage{multirow}
\usepackage{stackrel}
\usepackage{wrapfig}
\newcommand{\researchquestionslabel}[1]{%
    RQ#1:%
}
\newlist{researchquestions}{enumerate}{1}
\setlist[researchquestions]{%
    label=\researchquestionslabel{\arabic*},
    leftmargin=2.7em,
    labelwidth=*,
    labelsep=0.5em,
    itemindent=0pt,
    parsep=0pt
}

\definecolor{iccvblue}{rgb}{0.21,0.49,0.74}
\usepackage[pagebackref,breaklinks,colorlinks,allcolors=iccvblue]{hyperref}

\newcommand{\shortsection}[2][.]{\vspace{1mm}\noindent\textbf{#2#1}}

\def\paperID{15154} 
\def\confName{ICCV}
\def\confYear{2025}

\title{On the Robustness Tradeoff in Fine-Tuning}

\author{Kunyang Li, Jean-Charles Noirot Ferrand, Ryan Sheatsley, Blaine Hoak, \\ Yohan Beugin, Eric Pauley, Patrick McDaniel\\
University of Wisconsin-Madison\\
{\tt\small\{kli253, jcnf, sheatsley, bhoak, ybeugin, epauley, mcdaniel\}@cs.wisc.edu}
}

\begin{document}
\maketitle
\begin{abstract}
Fine-tuning has become the standard practice for adapting pre-trained models to downstream tasks. However, the impact on model robustness is not well understood. In this work, we characterize the robustness-accuracy trade-off in fine-tuning.  We evaluate the robustness and accuracy of fine-tuned models over 6 benchmark datasets and 7 different fine-tuning strategies. We observe a consistent trade-off between adversarial robustness and accuracy. Peripheral updates such as BitFit are more effective for simple tasks---over 75\% above the average measured by the area under the Pareto frontiers on CIFAR-10 and CIFAR-100. In contrast, fine-tuning information-heavy layers, such as attention layers via Compacter, achieves a better Pareto frontier on more complex tasks---57.5\% and 34.6\% above the average on Caltech-256 and CUB-200, respectively. Lastly, we observe that the robustness of fine-tuning against out-of-distribution data closely tracks accuracy. These insights emphasize the need for robustness-aware fine-tuning to ensure reliable real-world deployments.
\end{abstract}

\section{Introduction}
\label{sec:intro}
Pre-training and fine-tuning can efficiently transfer knowledge from upstream data to downstream tasks~\cite{brown_language_2020,tay_scale_2022}. Models can be fine-tuned in various ways~\cite{kumar_fine-tuning_2022}---full fine-tuning, linear probing (training the classification head), and parameter-efficient fine-tuning (PEFT)~\cite{hu_lora_2021,li_prefix-tuning_2021,chen_parameter-efficient_2023,edalati_krona_2022,guo_parameter-efficient_2021,houlsby_parameter-efficient_2019,le_fastfood_2014,lester_power_2021,liu_few-shot_2022,zaken_bitfit_2022}. Specifically, PEFT strategies selectively update parameters to achieve high accuracy on target tasks with significantly reduced computation and storage costs.

While fine-tuning optimizes efficiency and accuracy, its impact on model robustness remains underexplored--and in fact relatively unknown. Attackers actively generate adversarial examples~\cite{madry_towards_2019,papernot_limitations_2015} by adding crafted perturbations to cause model misclassification.   A model's ability to resist these samples, as well as perform well when presented with out-of-distribution (OOD) data, is called model robustness.

Prior studies on adversarial robustness~\cite{papernot_limitations_2015,madry_towards_2019,goodfellow_explaining_2015,carlini_towards_2017,croce_reliable_2020} focus on attacking models that are trained from scratch. Models learn to use high-dimensional features during training. Since test data have similar features to training data, models are able to achieve high accuracy at evaluation. However, attackers can generate adversarial examples to largely degrade accuracy by perturbing those features, which we refer to as highly predictive, non-robust features~\cite{ilyas_adversarial_2019,tsipras_robustness_2019} in this paper. In comparison, while attacks are targeted at downstream phenomena~\cite{chen_cartl_2021,jiang_robust_2020, chen_adversarial_2020, hua_initialization_2024, nguyen_adapters_2024}, two data phenomena are involved in pre-training and fine-tuning---upstream and downstream data, respectively. Here, a key question arises: \textit{how does adversarial robustness vary as the model is fine-tuned?} We hypothesize that features learned from pre-trained data are more robust against downstream attacks. As models fit to downstream phenomena, they learn non-robust features to gain accuracy while sacrificing adversarial robustness. Additionally, the fundamental trade-off between accuracy and adversarial robustness~\cite{tsipras_robustness_2019,li_trade-off_2023,zhang_theoretically_2019} may still exist but is potentially shifted. 

\begin{figure}[t]
    \centering
    \includegraphics[width=1.0\columnwidth]{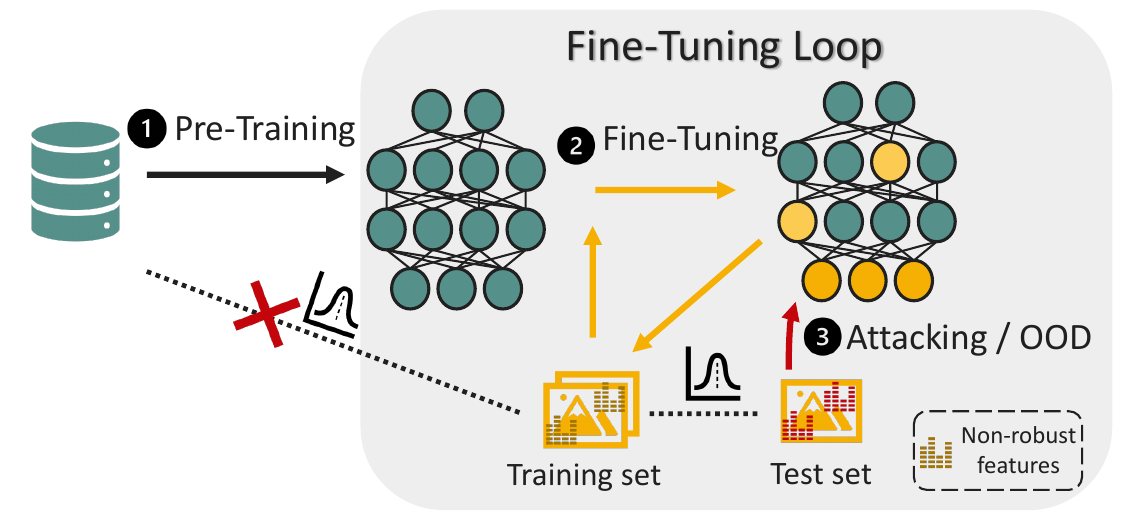}
    \caption{Continuous robustness evaluation during fine-tuning.}
    \label{fig:methodology}
\end{figure}

This work is the first to investigate deeply how robustness is impacted by fine-tuning. Here, we focus on three questions: (1) does the adversarial robustness-accuracy trade-off exist during fine-tuning? (2) how sensitive is this trade-off to different fine-tuning strategies and downstream data distributions? and (3) do findings on adversarial robustness generalize to OOD robustness? We begin by formally exploring the potential interaction between robustness and fine-tuning (\ie, the impact of mechanisms and data phenomena).  Thereafter, we evaluate trade-offs empirically by fine-tuning a test suite of models using a range of fine-tuning strategies (see \autoref{fig:methodology}). The experiments are performed over 231 models, 7 fine-tuning methods, and 6 benchmark datasets (5 for adversarial robustness and 1 with 6 domains for OOD robustness), resulting in approximately 2,100 adversarial and 2,000 OOD robustness assessments.

The evaluation finds: (1) a consistent adversarial robustness-accuracy trade-off early (within the first 3 epochs) in fine-tuning across all methods---initially, robustness improves together with accuracy, then it peaks and declines as fine-tuning continues; (2) the Pareto frontiers (\ie, optimal trade-off curves) are sensitive to fine-tuning methods and downstream task complexity---fine-tuning methods modifying intermediate information-intense layers, such as attention layers (\eg, Compacter~\cite{mahabadi_compacter_2021}), achieve better balances than those updating excessively (\eg, full fine-tuning) or only peripheral layers (\eg, linear probing, BitFit~\cite{zaken_bitfit_2022}); and (3) OOD robustness does not exhibit a trade-off with accuracy but remains relatively stable and closely tracks accuracy, suggesting that different underlying mechanisms drive robustness in security and safety contexts. These findings deepen our understanding of designing robustness-aware fine-tuning strategies facing different tasks and risks.  
\section{Background}\label{sec:background}
\subsection{Fine-tuning Strategies}
\shortsection{ViT backbone} The transformer architecture~\cite{vaswani_attention_2023} has become the state-of-the-art across many fields~\cite{dosovitskiy_image_2021,dong_speech-transformer_2018,brown_language_2020}. It typically serves as the backbone structure in the pre-training and fine-tuning paradigm~\cite{tay_scale_2022}, where general knowledge of pre-trained models is transferred to solve specific tasks through fine-tuning on (often small) downstream datasets. In this study, we focus on vision transformers (ViT), thus on images. Here, an input image is divided into fixed-size patches, flattened, and projected into embeddings. Then attention scores are calculated to determine the relationship between patches to capture global dependencies. The outputs are then passed through a feedforward network (FFN), in-between two layer normalizations (LN) for stability. 

\shortsection{Fine-tuning Methods} Full fine-tuning, which updates all model parameters, is widely used to transfer knowledge from pre-trained models to downstream tasks in computer vision~\cite{zhai_large-scale_2020,kumar_fine-tuning_2022}. While effective, it can be computationally expensive, especially for large-scale models. In contrast, linear probing, which fine-tunes only the final classification layer, is more efficient but often fails to match the performance of full fine-tuning~\cite{kumar_fine-tuning_2022}. To bridge this gap, parameter-efficient fine-tuning (PEFT) methods have been developed, aiming to achieve comparable or higher accuracy with fewer trainable parameters and reduced memory and storage overhead~\cite{he_parameter-efficient_2023,lialin_scaling_2024}. 

PEFT techniques primarily retain the pre-trained parameters while introducing a small set of trainable parameters to adapt the model to the downstream task. A general formulation can be expressed as: 
\begin{equation}\label{eq:peft}
    \hat y \leftarrow W_0x + \Delta Wx,
\end{equation}
where $W_0$ represents the frozen pre-trained weights, and $\Delta W$ denotes the task-specific learnable parameters introduced by a given PEFT method. The input $x$ is processed through the model, and the prediction $\hat y$ is computed based on all parameters. Several PEFT techniques have been adapted for ViTs from the original transformer models in NLP~\cite{adapterhub_adapterhub_2025}. LoRA~\cite{hu_lora_2021} reduces computational cost by factorizing weight updates into low-rank matrices within attention layers, effectively capturing task-specific adaptations. BitFit~\cite{zaken_bitfit_2022} takes a more selective approach by updating only bias terms, leaving all other weights unchanged. Adapter-based methods~\cite{houlsby_parameter-efficient_2019} introduce small trainable modules between transformer layers to inject task-specific information without modifying the backbone. Compacter~\cite{mahabadi_compacter_2021} further improves efficiency by using Kronecker-based parameterization within adapters, reducing the number of additional parameters needed. Finally, (IA)$^3$~\cite{liu_few-shot_2022} fine-tunes the model by learning per-layer multiplicative reweighting factors, modifying activations without directly changing pre-trained weights. A detailed visualization of how they are applied to a ViT block is shown in \autoref{fig:pefts}.

\subsection{Model Robustness}
Model robustness is crucial for evaluating the reliability of machine learning models under both adversarial manipulations and natural distribution shifts. Adversarial robustness~\cite{papernot_limitations_2015, goodfellow_explaining_2015,madry_towards_2019,carlini_towards_2017} focuses on a model's ability to defend against adversarial examples---carefully crafted perturbations that are imperceptible to humans but lead to incorrect model predictions. A widely used benchmark for security evaluation, projected gradient descent (PGD), iteratively modifies inputs to maximize model loss $L$ while constraining the perturbed image $x + \delta$ within a predefined norm-ball $\mathcal{B}$ of radius $\epsilon$ centered at the original input $x$ as shown in \autoref{eq:pgd}.

\begin{equation}\label{eq:pgd}
    x_{adv} = \text{argmax}_{x + \delta\in \mathcal{B}_\epsilon(x)} L(x + \delta, y)
\end{equation}

\noindent Beyond adversarial threats, models are also expected to demonstrate robustness to out-of-distribution (OOD) data, which is important to ensure reliable performance in real-world settings~\cite{peng_moment_2019,krizhevsky_imagenet_2017}. OOD shifts can vary in nature, from entirely novel objects absent in training data to more subtle domain variations, such as background changes or stylistic transformations (e.g., sketches versus real images). This study focuses on the latter, where the object of interest remains the same but appears in a different context.

\section{Methodology}

In this section, we construct two artifacts to study the relationship between robustness and fine-tuning.  We begin by extending an existing model of training to explore the potential impacts of fine-tuning on robustness, and then construct an evaluation framework to (a) decompose the space of fine-tuning strategies and (b) analyze how robustness varies with fine-tuning methods (reported in \autoref{sec:eval}).

\subsection{Modeling Robustness}\label{methodology:theoretical_model}
We establish the problem setup, based on prior work on model robustness~\cite{tsipras_robustness_2019}, to understand how fine-tuning affects robustness. Consider a binary downstream classification task, where labels are uniformly distributed---$y \stackrel{u.a.r}{\sim} \{-1, +1\}$. Each input $x$ consists of a feature, $x_1$, strongly-correlated to the corresponding ground-truth label (\ie, robust), and $d$ weakly-correlated (\ie, non-robust) features:
\begin{equation}\label{eq:weak_features}
    x_2, ..., x_{d+1} \stackrel{i.i.d}{\sim} \mathcal{N}(\eta y, 1).
\end{equation}
Here, $\eta$ represents the mean shift of the weakly correlated features, quantifying their predictive power. A larger $\eta$ implies that these features contribute more information toward classification, while a smaller $\eta$ means they are less distinguishable from noise. 

In fine-tuning, the classifier has frozen, pre-trained weights $w_0$ and adaptive weights $\Delta w$. Here, $k$ parameters are updated, where $k = \|\Delta w\|_0$, $d = \|w_0\|_0$, and $k \ll d$. A simple linear classifier is defined as: 
\begin{equation}\label{eq:ft_classifier}
    f_{FT}(x) := \text{sign}((w_0+\Delta w)^\top x),
\end{equation}
where
\begin{equation}\label{eq:weights}
    w_0 = [0, \frac{1}{d},...,\frac{1}{d}], \Delta w = [0,\frac{1}{d},..., 0,..., \frac{1}{d}].
\end{equation}
Then, we derive a lower bound on $\eta$ to analyze the robustness impact of fine-tuning:
\begin{equation}\label{eq:eta_lower_bound}
\begin{aligned}
    \Pr[f_{FT}(x) = y] &= \Pr[\text{sign}((w_0+\Delta w)^\top x)\cdot y > 0] \\
    &= \Pr[\text{sign}(\sum_{i=1}^d\frac{1}{d}x_i + \sum_{i=1}^k\frac{1}{d}x_i)\cdot y > 0] \\
    &= \Pr[\mathcal{N}(\frac{k+d}{d}\eta, \frac{k+d}{d^2}) > 0].
\end{aligned}
\end{equation}
Assuming a fine-tuned classifier achieves $99\%$ accuracy, we obtain (from the standard normal (Z) table):
\begin{equation}
    \eta \geq \frac{2.33}{\sqrt{k+d}} 
\end{equation}
This shows that the required correlation strength $\eta$ of non-robust features depends on both $k$ and $d$. For full fine-tuning (\ie, $k=d$), this simplifies to $\eta_{\text{full}} \geq \frac{2.33}{\sqrt{2d}}$, which relaxes its lower bound. Here, fine-tuning the entire model allows it to learn a larger number of non-robust features jointly to achieve high accuracy. But each feature is even less correlated to ground-truth, and thus, the model becomes more vulnerable. Additionally, the lower bound is tightened if the downstream task is simpler (\ie, smaller $d$), such as tasks with well-separated classes or fewer features. In this case, the model requires those non-robust features to have comparatively higher correlations. Thus, they are less susceptible to adversarial perturbations.

These preliminary results suggest that $k$, which is related to fine-tuning methods, and $d$, which is related to downstream task complexity, are connected to adversarial robustness. It motivates further exploration on measuring and studying those relationships.

\begin{figure*}[ht]
    \centering
    \includegraphics[width=0.9\textwidth]{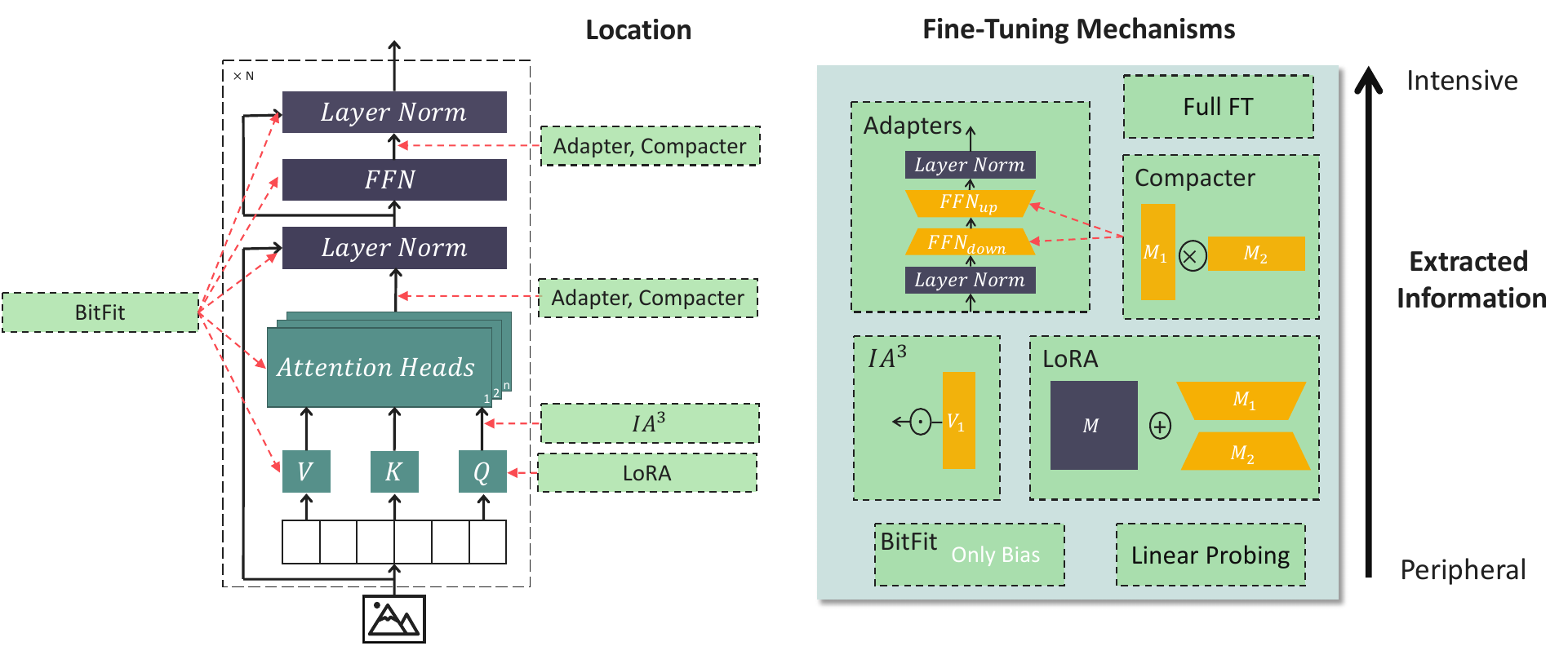}
    \caption{A graphical illustration of how 5 PEFT methods are applied to ViT (left) and a decomposition of PEFT mechanisms (right).}
    \label{fig:pefts}
\end{figure*}

\subsection{Measuring Robustness}

\subsubsection{Decomposition of PEFTs}\label{method:decomp_peft} 

To investigate the impact of fine-tuning on robustness, we select seven state-of-the-art fine-tuning strategies for our decomposition, including five PEFT methods that span all three main categories~\cite{lialin_scaling_2024}: addition-based (\ie, inserting new parameters: Adapter~\cite{houlsby_parameter-efficient_2019}, Compacter~\cite{mahabadi_compacter_2021}, and IA$^3$~\cite{liu_few-shot_2022}), reparametrization-based (\ie, decomposing into low-rank matrices: LoRA~\cite{hu_lora_2021}), and selection-based (\ie, modifying pre-trained weights: BitFit~\cite{zaken_bitfit_2022}), as well as full fine-tuning and linear probing. 

We decompose PEFT methods along two dimensions: a) the \textit{type} of information extracted from the pre-trained model and b) the \textit{mechanisms} used to fine-tune the extracted information. As illustrated in \autoref{fig:pefts}, we map out where PEFT strategies are applied within a ViT block and the underlying mechanisms they use. Since the addition of full fine-tuning and linear probing to the visualization is straightforward, we focus on decomposing PEFT methods here. 

The knowledge models gain during fine-tuning depends on what information PEFT methods extract from the pre-trained model. It includes the information type (\ie, model weights or representations) and its location. Here, weights correspond to static parameters in the pre-trained model layers, whereas intermediate representations are dynamic and dependent on input data. For example, LoRA~\cite{hu_lora_2021} explicitly modifies model weights by decomposing attention matrices into low-rank matrices, whereas (IA)$^3$~\cite{liu_few-shot_2022} introduces vectors to scale intermediate representations after the attention layer. Beyond distinguishing between weights and representations, we also examine where these modifications occur within the model. Different PEFT strategies target specific layers, such as attention weights, feed-forward networks (FFNs), or biases as shown in \autoref{fig:pefts}. 

Once information is extracted, PEFT strategies use specific mechanisms to update parameters. We identify three primary mechanisms used---(a) projection with neural layers, which introduces feed-forward layers or layer normalization to down-project and up-project intermediate representations; (b) matrix/vector computation, which applies matrix operations (\eg multiplication) to rescale extracted parameters; and (c) direct update, which directly uses backpropagation to update selected parameters.

These two dimensions remain consistent for each PEFT method across the $N$ blocks of a ViT. We provide a summary mapping table in \autoref{sec:app:decomp}. Our decomposition offers a new perspective on studying the fine-tuning space. This enables us to have a solid foundation to further investigate how and why different fine-tuning strategies may have different degrees of robustness. 

\subsubsection{Sensitivity Analysis}
Our framework, as shown in \autoref{fig:methodology}, systematically analyzes how robustness evolves throughout fine-tuning. As opposed to focusing on the final model state, it captures the dynamics of the trade-off between accuracy and robustness as the model is adapted. This is particularly important for fine-tuning. Here, the model transitions from a general to a specialized state with different numbers of robust and non-robust features learned. Encountering new data phenomena (\ie, step-level) and iterative updates (\ie, epoch-level) lead to changing degrees of the robustness-accuracy trade-off. 

Our approach builds on insights from overfitting studies. There, the model's test accuracy declines after prolonged training as it memorizes dataset-specific noises rather than generalizable patterns~\cite{ng_feature_2004}. Similarly, fine-tuned models may learn \textit{non-robust} features~\cite{ilyas_adversarial_2019,tsipras_robustness_2019} from downstream datasets to improve accuracy, but it degrades robustness. Here, our pipeline has two stages---(a) fine-tuning integration and (b) continuous evaluation. We first integrate PEFT modules by modifying the pre-trained model structure. Then, as the model is fine-tuned on downstream data, we use an adaptive tracking schedule to continuously evaluate model robustness and accuracy. 

Intuitively, robust and non-robust features learned from upstream and downstream data at different points influence model robustness and accuracy on target tasks. To capture the full variances during training, we monitor them at key update steps during fine-tuning. A major challenge here is to determine the optimal tracking frequency. While monitoring per epoch is standard, it is too coarse for classification tasks where fine-tuned models converge within a few epochs~\cite{hua_initialization_2024}. Early-stage changes of robustness are missed with sparse tracking. Instead, we track robustness at the granularity of backpropagation steps. This captures how the robustness-accuracy trade-off evolves in two ways: (a) when new downstream-specific data is introduced, revealing immediate shifts in robustness as the model encounters new data phenomena, and (b) as the model iteratively updates after seeing the entire dataset, showing longer-term trends in robustness and accuracy trade-offs.   

Furthermore, to balance efficiency while tracking at this granularity, we strategically sample model states at selected backpropagation intervals rather than at every step. Specifically, we increase tracking frequency (\ie, every 50-200 steps) during early training, when models rapidly adapt to new data, and decrease it (\ie, every 6,000 steps) in the later stage when performance stabilizes. This adaptive tracking approach ensures that we capture critical transitions in robustness while minimizing unnecessary overhead.
\section{Evaluation}\label{sec:eval}
Building upon our framework of fine-tuning integration and continual robustness evaluation, we empirically investigate how robustness changes during fine-tuning. Specifically, we seek to understand whether the shift from training a model from scratch to various fine-tuning strategies changes the robustness-accuracy trade-off. To this end, we address the following three key research questions:
\begin{researchquestions}
    \item Does the established trade-off between accuracy and adversarial robustness persist in fine-tuning? 
    \item How do different fine-tuning strategies and downstream task complexity affect the optimal trade-offs?
    \item Are the findings consistent with out-of-distribution (OOD) robustness?
\end{researchquestions}

\subsection{Experimental Design}\label{exp:setup}
To facilitate our experiments, we use ViT-Base model, pretrained on ImageNet-21k~\cite{ridnik_imagenet-21k_2021} (14 million images, 21,843 classes) at resolution 224x224 from HuggingFace~\cite{huggingface_googlevit-base-patch16-224-in21k_2025}, and AdapterHub~\cite{adapterhub_adapterhub_2025} v1.0.0 to integrate PEFT modules. Experiments are performed across 12 A100 GPUs with 40 GB of VRAM and CUDA version 11.7 or greater. Fine-tuning models with tracking adversarial and OOD robustness takes approximately 2,100 and 3,780 GPU hours, respectively. Differences among PEFT strategies are negligible---they all update 0.07\%-3.97\% of the pre-trained model. Our source code is publicly available.\footnote{\url{https://github.com/kyangl/robustness-finetuning}}

Since fine-tuning strategies are often used in data-limited regimes, we evaluate them on six representative datasets (10k-60k samples) with varying complexity (10 vs. 256 classes, fine- vs. course-grained classes): CIFAR10~\cite{recht_cifar-10_2018}, CIFAR100~\cite{krizhevsky_cifar-10_2009}, CalTech256~\cite{griffin_caltech_2022}, CUB200~\cite{wah_caltech-ucsd_2011}, StanfordDogs~\cite{khosla_novel_2011}, and DomainNet~\cite{peng_moment_2019}. Dataset details are in \autoref{sec:app:datasets}. To study scalability with data quantity, we also include Places365~\cite{zhou_places_2018} ($\sim$1.8M samples). However, as models converge to only $\sim$50\% clean accuracy~\cite{zhou_places_2018} and $\sim$5\% adversarial robustness, their weak overall performance makes it difficult to draw meaningful conclusions about robustness trends as discussed in \autoref{sec:app:data_scale}. Additionally, fine-tuning configurations follow standard CV practice~\cite{he_parameter-efficient_2023,hua_initialization_2024} (\autoref{tab:peft_hyperparameters}), with grid search used to select training hyperparameters (\autoref{sec:hyperparameters}). We also conduct an ablation study on learning rate and the number and location of trainable parameters, which can be found in \autoref{sec:app:ablation}.

\begin{table}[t]
    \centering
    \begin{tabular}{ l | l | l} 
    \hline
    \textbf{PEFTs} & \textbf{Configs} & \textbf{Values} \\ \hline
    \multirow{3}{*}{\makecell{Adapter \& \\ Compacter}} & reduction factor & 8\\
        & non linearity & gelu \\
        & locations     & multi-heads attn, $W_O$ \\
    \hline
    (IA)$^3$    &   locations   &  $W_K$, $W_V$, FFN\\ 
    LoRA        &   locations   &  $W_K$, $W_V$, $W_Q$, $W_O$ \\
    \bottomrule
    \end{tabular} 
    \caption{Standard configurations of PEFTs.}
\label{tab:peft_hyperparameters}
\end{table}

For adversarial robustness evaluations, we use the state-of-the-art attack algorithm PGD~\cite{madry_towards_2019} from TorchAttack~\cite{kim_torchattacks_2021} v3.5.1. Following standard practices~\cite{li_trade-off_2023}, we set the attack budget to $\epsilon = 1/255$, step size $\alpha = 0.25/255$, and the number of steps to $15$. Adversarial examples are generated from the test sets of each downstream dataset. Given the computational cost of attacks, we follow a structured evaluation tracking schedule: (1) in early fine-tuning ($0$-$700$ steps), we evaluate robustness and accuracy every $50$ steps to capture robustness changes; (2) between $700-3,000$ steps, evaluations occur every $1,000$ steps; and (3) beyond $3,000$ steps, we evaluate every $6,000$ steps. 

For OOD robustness, we evaluate the model's ability to generalize across distribution shifts using DomainNet~\cite{peng_moment_2019}. Here, the model is fine-tuned on a single domain and tested on other unseen domains. Due to the large number of data with higher computational overhead, OOD evaluations are conducted at a coarser granularity: (1) every $200$ steps for the first $1,000$ fine-tuning steps; (2) every $2,000$ steps from $1,000$ to $3,000$ steps; (3) every $4,000$ steps from $3,000$ to $10,000$ steps; (4) every $6,000$ steps from $10,000$ to $30,000$ steps; and (5) every $20,000$ beyond $30,000$ steps.  

\subsection{Accuracy and Robustness Trade-off}
Prior studies attribute the trade-off between adversarial robustness and accuracy to models that rely on a large number of non-robust yet highly predictive features~\cite{ilyas_adversarial_2019,tsipras_robustness_2019}. This claim is based on the assumption that the training data and the data used to generate adversarial examples share the same distribution. However, this assumption breaks in fine-tuning. Here, distinct downstream datasets with different inter-class/domain separation, image resolution, and similarities with upstream data are involved during training. This shift raises a crucial question: does the trade-off phenomenon still persist in fine-tuning and how? 

\begin{figure}[t]
    \centering
\includegraphics[width=1\columnwidth]{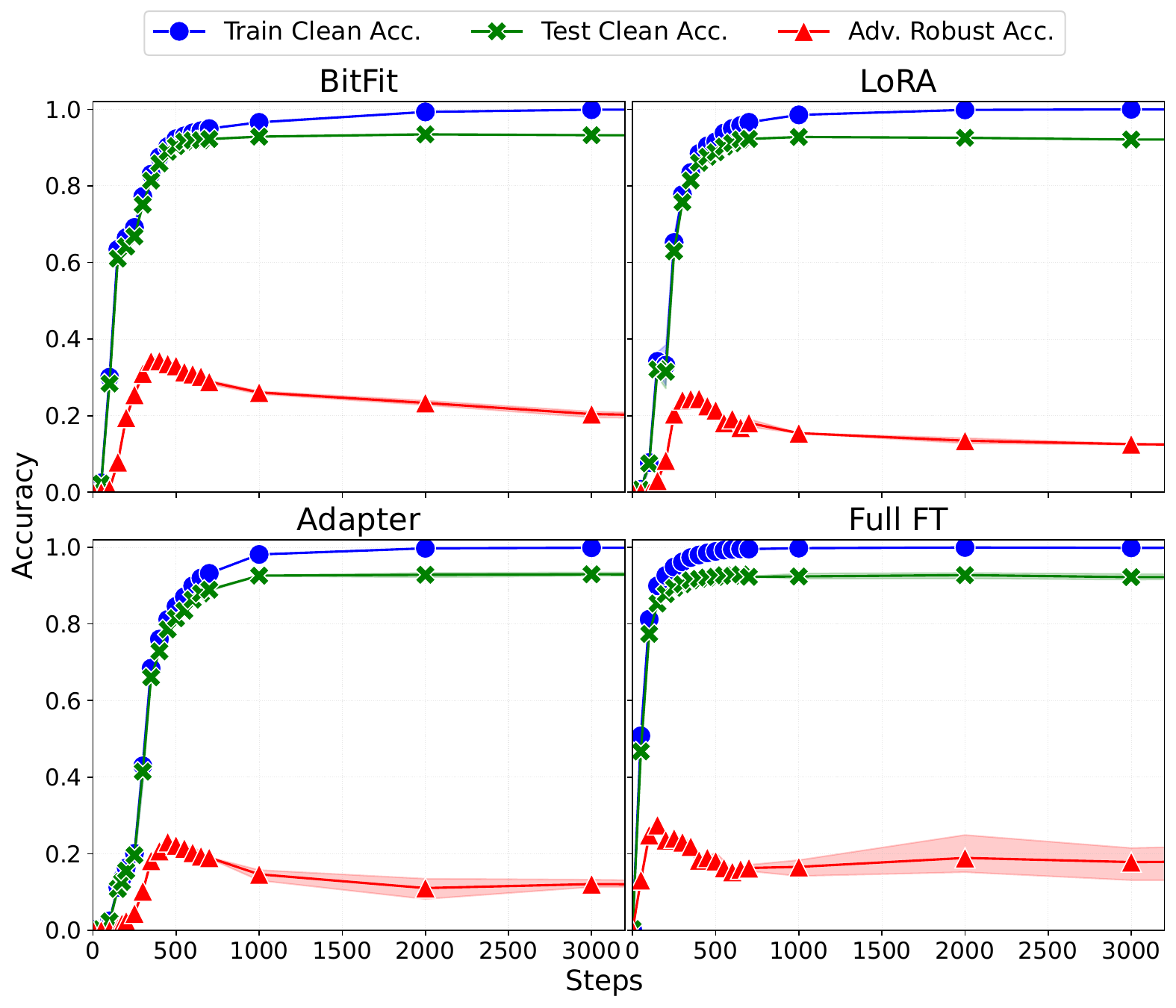}
    \caption{Continuous evaluation of training accuracy (blue), test accuracy (green), and adversarial robustness (red) across backpropagation steps (truncated at 3000 steps) on Caltech256.}
    \label{fig:training_curves}
\end{figure}

To explore this, we measure the variances of robustness and accuracy during fine-tuning for 7 fine-tuning methods and 5 datasets. \autoref{fig:training_curves} shows results on \texttt{Caltech256}, where models trained with BitFit, LoRA, Adapter, and full fine-tuning all exhibit rapid improvements in standard accuracy, reaching $\approx 90\%$ within $1,000$ steps. Meanwhile, adversarial robustness follows a different trajectory: it initially increases, then reaches $\approx 25\%$ at around step $400$, and finally steadily declines to $\approx 10\%$ at convergence. Specifically, PEFT methods that fine-tune parameters in or around attention layers (\eg, LoRA, Adapter) demonstrate a slightly more gradual decline in robustness, suggesting that they preserve robustness better than full fine-tuning or bias-only tuning. We further investigate differences among fine-tuning strategies in the next section.

The distinct trends of accuracy and robustness here highlight that models learn features for different purposes at different stages. Early fine-tuning steps improve both robustness and accuracy by leveraging pre-trained representations. We attribute this to the model's ability to effectively adapt the \textit{randomly initialized} trainable parameters to downstream tasks. Then, as the model is increasingly fitted to the downstream data, it begins to exploit predictive but non-robust features to gain accuracy while sacrificing robustness. The findings strongly confirm the persistence of the trade-off in fine-tuning and imply that learning robust features with defense mechanisms may decrease accuracy. 

\begin{figure}[h]
    \centering
    \includegraphics[width=1\linewidth]{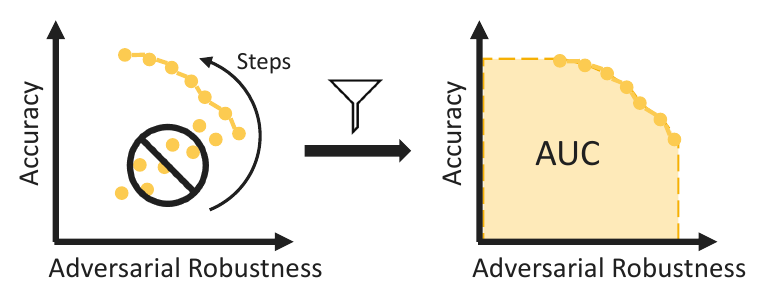}
    \caption{
    Pareto frontiers are extracted by filtering out the suboptimal points in the trade-off space. The AUC is then computed by extending the two endpoints and integrating the enclosed area.}
    \label{fig:auc}
\end{figure}

\subsection{Pareto Frontiers in the Trade-off Space}\label{sec:eval_pareto_curves}
After analyzing the \textit{consistent} adversarial robustness-accuracy trade-off, we further ask how \textit{sensitive} the trade-off is to downstream phenomena and fine-tuning methods.

\begin{figure}[t]
    \centering
\includegraphics[width=1\columnwidth]{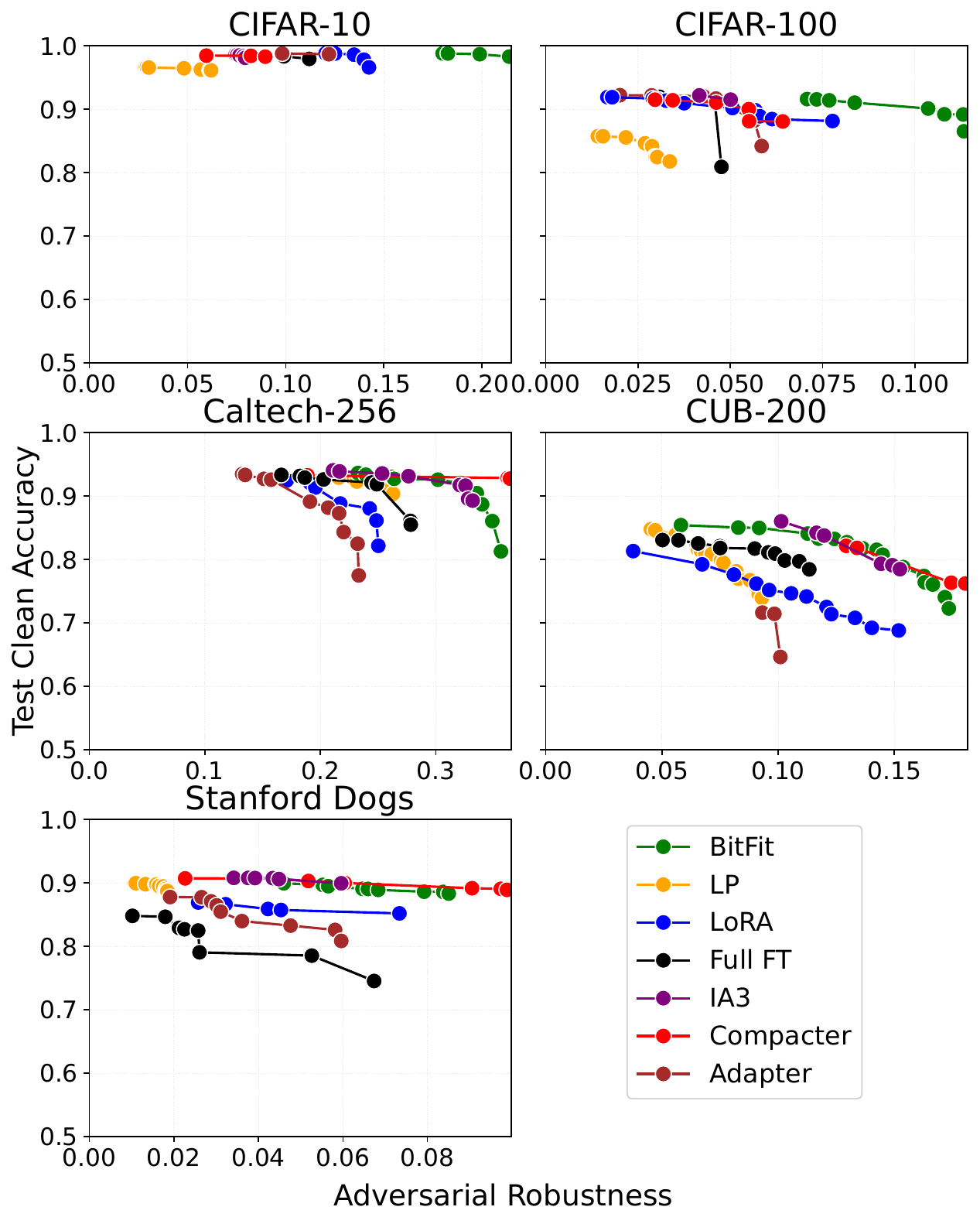}
    \caption{Pareto frontiers of the trade-off between accuracy and robustness on five downstream datasets.}
    \label{fig:pareto_front_curves}
    \vspace{-1.5em}
\end{figure}

\shortsection{Across downstream distributions} First, we extract Pareto frontiers by identifying points in the trade-off space that no other point has higher accuracy and robustness simultaneously. As shown in \autoref{fig:auc}, these frontiers represent the optimal trade-offs achieved during fine-tuning. \autoref{fig:pareto_front_curves} illustrates the results of the fine-tuning methods on the five datasets. It shows significant differences across downstream distributions. CIFAR-10 exhibits the flattest Pareto frontiers (\ie, the most gradual trade-offs). Adversarial robustness remains relatively stable as accuracy improves in the last $2\%$ before convergence (leftmost points of each frontier). Similarly, CIFAR-100 shows less gradual trade-offs, characterized by smaller gradients (in absolute values) of the frontiers. In comparison, the Pareto frontiers are steeper, indicating a more evident trade-off, for Caltech256, CUB200, and Stanford Dogs. Here, the robustness peaks (the rightmost points of each curve) and then sharply declines as accuracy approaches the final $10\%$ before convergence.

We attribute this variation to task complexity. CIFAR-10 consists of $10$ classes that fully overlap with the pre-trained data. This enables a smoother adaptation. In contrast, CUB-200 requires distinguishing 200 bird species, all falling under the broad ``Bird" category of the upstream data. The complexity with smaller inter-class separation forces the model to learn finer details, increasing its dependence on fragile, non-robust features. In summary, greater downstream task complexity with less similarity with the upstream phenomena leads to steeper Pareto frontiers. 

\shortsection{Across fine-tuning strategies} 
Furthermore, to quantify the quality of the trade-off across fine-tuning strategies, we compute a simple scalar---area under the Pareto frontiers \cite{little_fairness_2022}---as our metric. Since Pareto frontiers capture the optimal balances that are achievable by each fine-tuning strategy, we aggregate all ``suboptimal" points in that trade-off space by extending the two end points of the Pareto frontier and integrating the enclosed area, as shown in \autoref{fig:auc}. We refer to this metric as AUC in the paper. Here, a larger value indicates a better balance between robustness and accuracy.

BitFit achieves the highest AUC for CIFAR10 (75\% above the average) and CIFAR100 ($81.5\%$), while Compacter outperforms others on Caltech256 by 57.5\%, Stanford Dogs by 24\%, and CUB200 by 34.6\%. This suggests that BitFit excels on simpler datasets, where modifying only bias terms efficiently adapts pre-trained knowledge while retaining robustness. In contrast, Compacter is better for complex tasks, where its low-rank reparameterization in intermediate layers balances adaptation and robustness inheritance better. Notably, linear probing (LP) and full fine-tuning (Full FT) underperform across all datasets, with the lowest AUCs for CIFAR100 (47.5\% and 20\% lower than the average, respectively) and CUB200 (28.2\% and 19.2\%, respectively). As shown in \autoref{fig:pareto_front_curves}, LP and Full FT result in shorter Pareto frontiers, failing to achieve either high robustness or high accuracy compared to others. Fine-tuning all parameters destabilizes robustness, while freezing the entire network except for the classification layer limits adaptation, both resulting in suboptimal trade-offs. 

The observed trends align well with our preliminary results (\autoref{methodology:theoretical_model}). Here, the fine-tuning methods, corresponding to $k$, and downstream data phenomena, corresponding to $d$, lead to different degrees of model robustness. Furthermore, we attribute the different results of different fine-tuning methods to where (\ie, information location) and how (\ie, underlying mechanisms) the models are fine-tuned, as described in \autoref{method:decomp_peft}. For example, BitFit and LP are applied at the periphery of the model, adjusting minimal information, while LoRA, Adapter, and Compacter update deeper layers (\eg, attention mechanisms), introducing information-intense updates. Additionally, full fine-tuning adapts all parameters, leading to a more suboptimal trade-off. Here, methods that adapt intermediate representations maintain better robustness while improving accuracy, whereas peripheral or excessive updates largely degrade the balance between robustness and accuracy.

\begin{table}[t]
    \centering 
    \begin{tabular}{lccccc}
    \toprule
    {} &  C10 &  C100  &  Cal &  CUB    &  Dogs \\
    \midrule
    BitFit      &   \textbf{0.21}  &    \textbf{0.10}    &      0.33           &  0.14 &        0.08            \\
    Adapter    &   0.12        &    0.05                &      0.21           &  0.07   &        0.05         \\   
    LoRA        &   0.14       &    0.07                &      0.23           &  0.12   &        0.06         \\
    Compacter    &   0.09        &    0.06             &      \textbf{0.34}  &  \textbf{0.15}   &        \textbf{0.09}  \\
    IA3            &   0.08      &    0.05             &      0.31           &  0.13    &        0.05          \\
    LP          &   0.06      &    0.03               &      0.24           &  0.08 &        0.02           \textbf{} \\
    Full FT       &   0.11      &    0.04                &      0.26           &  0.09  &        0.05        \\
    \bottomrule
    \end{tabular}
    \caption{Area under the curve (AUC) of the Pareto frontiers.}
    \label{tab:pareto_auc}
\end{table}

\subsection{On Out-of-Distribution Robustness}
Here, we address our third research question---does the robustness-accuracy trade-off observed in adversarial settings extend to real-world out-of-distribution (OOD) scenarios? Unlike adversarial robustness, where non-robust features exploited by attacks contribute to the trade-off, OOD robustness depends on a model's ability to generalize beyond its training distribution. This fundamental difference suggests that fine-tuning strategies may show different behaviors in OOD settings. 

\begin{figure}[t]
    \centering
\includegraphics[width=1\columnwidth]{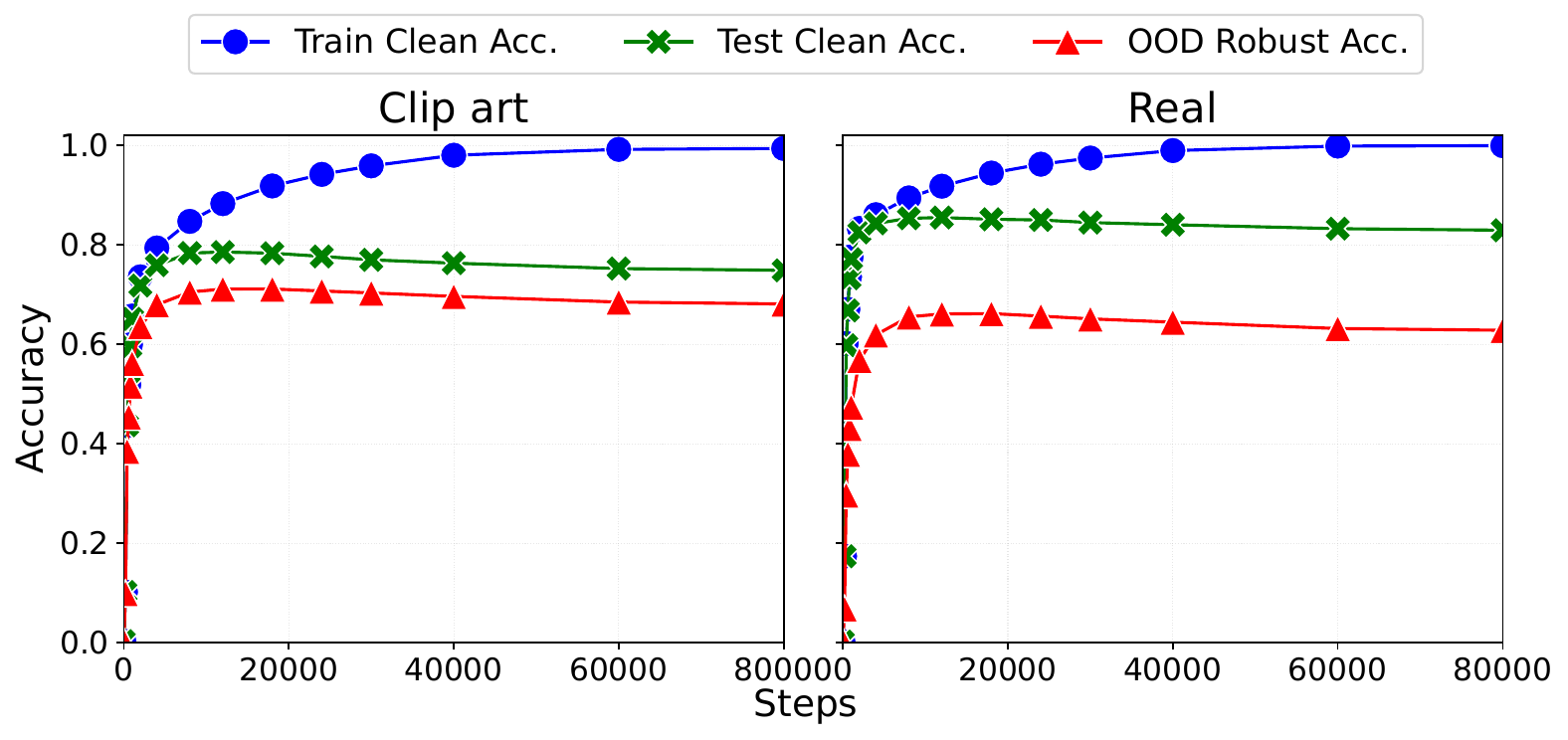}
  \caption{Continuous evaluation on training accuracy (blue), test accuracy (green), and OOD robustness (red) across backpropagation steps on clip art and real images from DomainNet.}\label{fig:ood_training_curves}
\end{figure}

Similarly, we track OOD robustness during fine-tuning and compare it to in-domain (test and training) accuracy, as shown in \autoref{fig:ood_training_curves}. In comparison to adversarial robustness, which often deteriorates after peaking, OOD robustness \textit{plateaus} after the initial improvement at a lower level than standard accuracy. Both in-domain and OOD robustness slightly decline after convergence, likely due to overfitting rather than the accuracy-robustness conflict in adversarial settings. 
This behavior highlights the different mechanisms behind adversarial and OOD robustness. While adversarial robustness is affected by low-level, non-robust features, OOD robustness depends on transferable features that are less sensitive to fine-tuning-induced degradation. 

\begin{figure}[t]
    \centering
    \includegraphics[width=1\columnwidth]{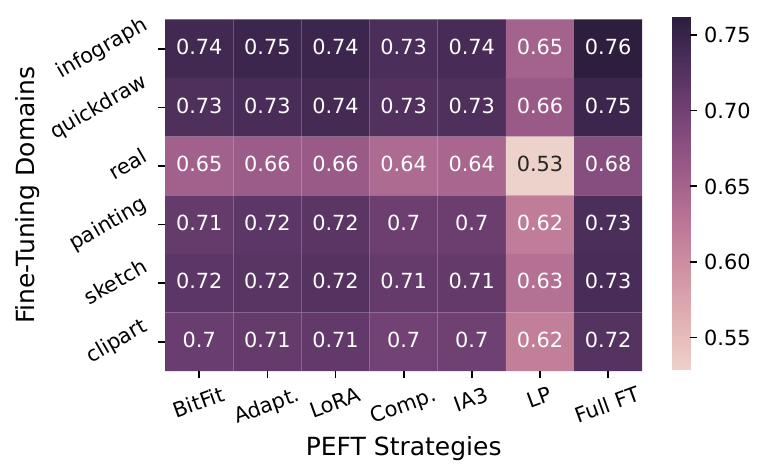}
    \caption{Heatmap representation of peak OOD robustness.}
    \label{fig:ood_peak_robustness}
\end{figure}

As shown in \autoref{fig:ood_peak_robustness}, we further analyze OOD robustness across the seven fine-tuning strategies and six training domains. Linear probing consistently yields the lowest OOD robustness ($61\%\pm5\%$), while full fine-tuning achieves the highest ($73\%\pm2\%$). In addition, models fine-tuned on the ``real" domain, which is closest to the pre-training distribution, exhibit lower OOD robustness ($64\%\pm5\%$) compared to more shifted domains such as ``infograph" ($73\%\pm4\%$) and ``quickdraw" ($72\%\pm3\%$). Interestingly, OOD robustness stays stable across fine-tuning strategies, excluding linear probing. This suggests that fine-tuning methods support model generalization to other domains similarly. The results reinforce that OOD robustness is primarily dependent on domain shifts and the extent of parameter updates (\eg linear probing vs. full fine-tuning), rather than the specific underlying mechanisms. 

\section{Related Work}\label{sec:related_work}
The fundamental trade-off between robustness and accuracy has been extensively studied~\cite{tsipras_robustness_2019,ilyas_adversarial_2019,zhang_theoretically_2019}. \citet{tsipras_robustness_2019} argue that this trade-off arises from the data distribution, where feature representations learned for high clean accuracy often rely on weakly-related (to ground-truth label) features that degrade under adversarial conditions. \citet{ilyas_adversarial_2019} further highlight that adversarial vulnerability is a consequence of models exploiting highly predictive yet non-robust features. \citet{zhang_theoretically_2019} propose a theoretically justified adversarial training framework to mitigate this trade-off but acknowledge the inherent tension between robustness and accuracy. However, these studies primarily focus on (a) full model training and (b) attacking with data (test set) similar to the training data, whereas our work examines this trade-off within the context of pre-training and fine-tuning, considering various fine-tuning methods and distinct upstream-downstream data phenomena shifts.

Furthermore, with the rise of fine-tuning, robustness considerations of them have become increasingly important. Prior research has explored robustness from a broader perspective, focusing on studying adversarial robustness either during (a) pre-training robustness~\cite{chen_cartl_2021,jiang_robust_2020,chen_adversarial_2020} or (b) fine-tuning~\cite{xu_autolora_2023,liu_twins_2023,jeddi_simple_2020}. Specifically, these studies primarily investigate full fine-tuning or linear probing, without considering various PEFT strategies. Recent studies have attempted to improve PEFT robustness. \citet{hua_initialization_2024} propose a robustness-aware initialization strategy. AdapterMixup~\cite{nguyen_adapters_2024} integrates adversarial training with adapters using mixup. CLAT~\cite{gopal_criticality_2024} studies layer-level robustness under adversarial training, while we focus on method-level evaluation across fine-tuning strategies. Additionally, they study the final model state instead of investigating with a finer granularity on how the robustness-accuracy tension changes fundamentally throughout fine-tuning before introducing robustness-improving mechanisms.
\section{Discussion \& Limitations} \label{sec:discussion}
\noindent \textbf{Non-robust features \& Robustness.} As adversarial robustness is closely tied to non-robust features from downstream data, one natural question arises: \textit{are there non-robust features from upstream data?} The ``robustness" of features here is relative. Since the goal is to apply models to solve downstream tasks, we only consider model accuracy and robustness on downstream phenomena. Thus, non-robust features specific to upstream data may be robust to adversarial perturbations targeted at downstream data. However, it is convoluted to distinguish \textit{when} the model learns \textit{what} features from different data phenomena when models can also learn them simultaneously. One potential way to further investigate this is to track robustness using adversarial examples generated on both upstream and downstream data. 

\noindent \textbf{Limitations \& future work.} Our study focuses on understanding robustness dynamics of SOTA PEFT strategies before applying defense mechanisms. Due to runtime constraints in continuous evaluation, we use PGD, which has been shown to reliably estimate robustness for clean models~\cite{croce_reliable_2020}. Future work can extend our framework to additional model types (\eg, CNN-ViT hybrid models, adversarially-trained models) and broader threat scenarios (\eg, black-box, adaptive, and data corruption attacks).

\section{Conclusion}
Our study systematically examines the trade-off between robustness and accuracy in the fine-tuning paradigm, revealing that adversarial robustness initially improves but declines as fine-tuning continues. Across 231 fine-tuned models with 7 state-of-the-art fine-tuning strategies and our independent security and safety evaluation framework, we find that the balance of the trade-off varies with fine-tuning strategies, downstream data complexity, and its similarity to pre-training data. Specifically, methods updating attention-related layers (\eg, LoRA, Compacter) tend to better balance robustness and accuracy, while simpler adaptation techniques (\eg, BitFit) achieve higher robustness peaks but degrade faster. However, this trade-off does not extend to safety (OOD) robustness, which instead depends on the model's ability to generalize across domains. These findings suggest that the design of robustness-aware fine-tuning strategies should consider both adversarial and OOD robustness independently.

\noindent \textbf{Acknowledgements.} We would like to thank Kassem Fawaz and Rahul Chatterjee for their helpful comments on earlier iterations of this work. This material is based upon work supported by the National Science Foundation under Grant No. CNS-2343611 and by the U.S. Army Research Office under MURI grant No. W911NF-21-1-0317. Any opinions, findings, and conclusions or recommendations expressed in this material are those of the author(s) and do not necessarily reflect the views of the National Science Foundation or the U.S. Army Research Office. The U.S. Government is authorized to reproduce and distribute reprints for government purposes notwithstanding any copyright notation hereon. 

{
    \small
    \bibliographystyle{ieeenat_fullname}
    \bibliography{main}

\begin{thebibliography}{52}
\providecommand{\natexlab}[1]{#1}
\providecommand{\url}[1]{\texttt{#1}}
\expandafter\ifx\csname urlstyle\endcsname\relax
  \providecommand{\doi}[1]{doi: #1}\else
  \providecommand{\doi}{doi: \begingroup \urlstyle{rm}\Url}\fi

\bibitem[AdapterHub(2025)]{adapterhub_adapterhub_2025}
AdapterHub.
\newblock {AdapterHub} {Documentation} — {AdapterHub} documentation, 2025.

\bibitem[Brown et~al.(2020)Brown, Mann, Ryder, Subbiah, Kaplan, Dhariwal, Neelakantan, Shyam, Sastry, Askell, Agarwal, Herbert-Voss, Krueger, Henighan, Child, Ramesh, Ziegler, Wu, Winter, Hesse, Chen, Sigler, Litwin, Gray, Chess, Clark, Berner, McCandlish, Radford, Sutskever, and Amodei]{brown_language_2020}
Tom~B. Brown, Benjamin Mann, Nick Ryder, Melanie Subbiah, Jared Kaplan, Prafulla Dhariwal, Arvind Neelakantan, Pranav Shyam, Girish Sastry, Amanda Askell, Sandhini Agarwal, Ariel Herbert-Voss, Gretchen Krueger, Tom Henighan, Rewon Child, Aditya Ramesh, Daniel~M. Ziegler, Jeffrey Wu, Clemens Winter, Christopher Hesse, Mark Chen, Eric Sigler, Mateusz Litwin, Scott Gray, Benjamin Chess, Jack Clark, Christopher Berner, Sam McCandlish, Alec Radford, Ilya Sutskever, and Dario Amodei.
\newblock Language {Models} are {Few}-{Shot} {Learners}, 2020.
\newblock arXiv:2005.14165 [cs].

\bibitem[Carlini and Wagner(2017)]{carlini_towards_2017}
Nicholas Carlini and David Wagner.
\newblock Towards {Evaluating} the {Robustness} of {Neural} {Networks}, 2017.
\newblock arXiv:1608.04644 [cs].

\bibitem[Chen et~al.(2021)Chen, Hu, Wang, Li, Wang, Shen, and Li]{chen_cartl_2021}
Dian Chen, Hongxin Hu, Qian Wang, Yinli Li, Cong Wang, Chao Shen, and Qi Li.
\newblock {CARTL}: {Cooperative} {Adversarially}-{Robust} {Transfer} {Learning}, 2021.
\newblock arXiv:2106.06667 [cs].

\bibitem[Chen et~al.(2023)Chen, Zhang, Shi, Li, Smola, and Yang]{chen_parameter-efficient_2023}
Jiaao Chen, Aston Zhang, Xingjian Shi, Mu Li, Alex Smola, and Diyi Yang.
\newblock Parameter-{Efficient} {Fine}-{Tuning} {Design} {Spaces}, 2023.
\newblock arXiv:2301.01821 [cs].

\bibitem[Chen et~al.(2020)Chen, Liu, Chang, Cheng, Amini, and Wang]{chen_adversarial_2020}
Tianlong Chen, Sijia Liu, Shiyu Chang, Yu Cheng, Lisa Amini, and Zhangyang Wang.
\newblock Adversarial {Robustness}: {From} {Self}-{Supervised} {Pre}-{Training} to {Fine}-{Tuning}, 2020.
\newblock arXiv:2003.12862 [cs].

\bibitem[Croce and Hein(2020)]{croce_reliable_2020}
Francesco Croce and Matthias Hein.
\newblock Reliable evaluation of adversarial robustness with an ensemble of diverse parameter-free attacks, 2020.
\newblock arXiv:2003.01690 [cs].

\bibitem[Dong et~al.(2018)Dong, Xu, and Xu]{dong_speech-transformer_2018}
Linhao Dong, Shuang Xu, and Bo Xu.
\newblock Speech-{Transformer}: {A} {No}-{Recurrence} {Sequence}-to-{Sequence} {Model} for {Speech} {Recognition}.
\newblock In \emph{2018 {IEEE} {International} {Conference} on {Acoustics}, {Speech} and {Signal} {Processing} ({ICASSP})}, pages 5884--5888, 2018.
\newblock ISSN: 2379-190X.

\bibitem[Dosovitskiy et~al.(2021)Dosovitskiy, Beyer, Kolesnikov, Weissenborn, Zhai, Unterthiner, Dehghani, Minderer, Heigold, Gelly, Uszkoreit, and Houlsby]{dosovitskiy_image_2021}
Alexey Dosovitskiy, Lucas Beyer, Alexander Kolesnikov, Dirk Weissenborn, Xiaohua Zhai, Thomas Unterthiner, Mostafa Dehghani, Matthias Minderer, Georg Heigold, Sylvain Gelly, Jakob Uszkoreit, and Neil Houlsby.
\newblock An {Image} is {Worth} 16x16 {Words}: {Transformers} for {Image} {Recognition} at {Scale}, 2021.
\newblock arXiv:2010.11929 [cs].

\bibitem[Edalati et~al.(2022)Edalati, Tahaei, Kobyzev, Nia, Clark, and Rezagholizadeh]{edalati_krona_2022}
Ali Edalati, Marzieh Tahaei, Ivan Kobyzev, Vahid~Partovi Nia, James~J. Clark, and Mehdi Rezagholizadeh.
\newblock {KronA}: {Parameter} {Efficient} {Tuning} with {Kronecker} {Adapter}, 2022.
\newblock arXiv:2212.10650 [cs].

\bibitem[Goodfellow et~al.(2015)Goodfellow, Shlens, and Szegedy]{goodfellow_explaining_2015}
Ian~J. Goodfellow, Jonathon Shlens, and Christian Szegedy.
\newblock Explaining and {Harnessing} {Adversarial} {Examples}, 2015.
\newblock arXiv:1412.6572 [stat].

\bibitem[Gopal et~al.(2024)Gopal, Yang, Zhang, Horton, and Chen]{gopal_criticality_2024}
Bhavna Gopal, Huanrui Yang, Jingyang Zhang, Mark Horton, and Yiran Chen.
\newblock Criticality {Leveraged} {Adversarial} {Training} ({CLAT}) for {Boosted} {Performance} via {Parameter} {Efficiency}, 2024.
\newblock arXiv:2408.10204 [cs].

\bibitem[Griffin et~al.(2022)Griffin, Holub, and Perona]{griffin_caltech_2022}
Gregory Griffin, Alex Holub, and Pietro Perona.
\newblock Caltech 256, 2022.

\bibitem[Guo et~al.(2021)Guo, Rush, and Kim]{guo_parameter-efficient_2021}
Demi Guo, Alexander~M. Rush, and Yoon Kim.
\newblock Parameter-{Efficient} {Transfer} {Learning} with {Diff} {Pruning}, 2021.
\newblock arXiv:2012.07463 [cs].

\bibitem[He et~al.(2023)He, Li, Zhang, Yang, and Wang]{he_parameter-efficient_2023}
Xuehai He, Chunyuan Li, Pengchuan Zhang, Jianwei Yang, and Xin~Eric Wang.
\newblock Parameter-efficient {Model} {Adaptation} for {Vision} {Transformers}, 2023.
\newblock arXiv:2203.16329 [cs].

\bibitem[Houlsby et~al.(2019)Houlsby, Giurgiu, Jastrzebski, Morrone, Laroussilhe, Gesmundo, Attariyan, and Gelly]{houlsby_parameter-efficient_2019}
Neil Houlsby, Andrei Giurgiu, Stanislaw Jastrzebski, Bruna Morrone, Quentin~de Laroussilhe, Andrea Gesmundo, Mona Attariyan, and Sylvain Gelly.
\newblock Parameter-{Efficient} {Transfer} {Learning} for {NLP}, 2019.
\newblock arXiv:1902.00751 [cs].

\bibitem[Hu et~al.(2021)Hu, Shen, Wallis, Allen-Zhu, Li, Wang, Wang, and Chen]{hu_lora_2021}
Edward~J. Hu, Yelong Shen, Phillip Wallis, Zeyuan Allen-Zhu, Yuanzhi Li, Shean Wang, Lu Wang, and Weizhu Chen.
\newblock {LoRA}: {Low}-{Rank} {Adaptation} of {Large} {Language} {Models}, 2021.
\newblock arXiv:2106.09685 [cs].

\bibitem[Hua et~al.(2024)Hua, Gu, Xue, Carlini, Wong, and Qin]{hua_initialization_2024}
Andong Hua, Jindong Gu, Zhiyu Xue, Nicholas Carlini, Eric Wong, and Yao Qin.
\newblock Initialization {Matters} for {Adversarial} {Transfer} {Learning}, 2024.
\newblock arXiv:2312.05716 [cs].

\bibitem[HuggingFace(2025)]{huggingface_googlevit-base-patch16-224-in21k_2025}
HuggingFace.
\newblock google/vit-base-patch16-224-in21k · {Hugging} {Face}, 2025.

\bibitem[Ilyas et~al.(2019)Ilyas, Santurkar, Tsipras, Engstrom, Tran, and Madry]{ilyas_adversarial_2019}
Andrew Ilyas, Shibani Santurkar, Dimitris Tsipras, Logan Engstrom, Brandon Tran, and Aleksander Madry.
\newblock Adversarial {Examples} {Are} {Not} {Bugs}, {They} {Are} {Features}, 2019.
\newblock arXiv:1905.02175 [stat].

\bibitem[Jeddi et~al.(2020)Jeddi, Shafiee, and Wong]{jeddi_simple_2020}
Ahmadreza Jeddi, Mohammad~Javad Shafiee, and Alexander Wong.
\newblock A {Simple} {Fine}-tuning {Is} {All} {You} {Need}: {Towards} {Robust} {Deep} {Learning} {Via} {Adversarial} {Fine}-tuning, 2020.
\newblock arXiv:2012.13628 [cs].

\bibitem[Jiang et~al.(2020)Jiang, Chen, Chen, and Wang]{jiang_robust_2020}
Ziyu Jiang, Tianlong Chen, Ting Chen, and Zhangyang Wang.
\newblock Robust {Pre}-{Training} by {Adversarial} {Contrastive} {Learning}.
\newblock In \emph{Advances in {Neural} {Information} {Processing} {Systems}}, pages 16199--16210. Curran Associates, Inc., 2020.

\bibitem[Khosla et~al.(2011)Khosla, Jayadevaprakash, Yao, and Li]{khosla_novel_2011}
Aditya Khosla, Nityananda Jayadevaprakash, Bangpeng Yao, and Fei-Fei Li.
\newblock Novel {Dataset} for {Fine}-{Grained} {Image} {Categorization}: {Stanford} {Dogs}.
\newblock In \emph{Workshop on {Fine}-{Grained} {Visual} {Categorization} ({FGVC}), {IEEE} {Conference} on {Computer} {Vision} and {Pattern} {Recognition} ({CVPR})}, 2011.

\bibitem[Kim(2021)]{kim_torchattacks_2021}
Hoki Kim.
\newblock Torchattacks: {A} {PyTorch} {Repository} for {Adversarial} {Attacks}, 2021.
\newblock arXiv:2010.01950 [cs].

\bibitem[Krizhevsky(2009)]{krizhevsky_cifar-10_2009}
Alex Krizhevsky.
\newblock {CIFAR}-10 and {CIFAR}-100 datasets, 2009.

\bibitem[Krizhevsky et~al.(2017)Krizhevsky, Sutskever, and Hinton]{krizhevsky_imagenet_2017}
Alex Krizhevsky, Ilya Sutskever, and Geoffrey~E. Hinton.
\newblock {ImageNet} classification with deep convolutional neural networks.
\newblock \emph{Commun. ACM}, 60\penalty0 (6):\penalty0 84--90, 2017.

\bibitem[Kumar et~al.(2022)Kumar, Raghunathan, Jones, Ma, and Liang]{kumar_fine-tuning_2022}
Ananya Kumar, Aditi Raghunathan, Robbie Jones, Tengyu Ma, and Percy Liang.
\newblock Fine-{Tuning} can {Distort} {Pretrained} {Features} and {Underperform} {Out}-of-{Distribution}, 2022.
\newblock arXiv:2202.10054 [cs].

\bibitem[Le et~al.(2014)Le, Sarlos, and Smola]{le_fastfood_2014}
Quoc~Viet Le, Tamas Sarlos, and Alexander~Johannes Smola.
\newblock Fastfood: {Approximate} {Kernel} {Expansions} in {Loglinear} {Time}, 2014.
\newblock arXiv:1408.3060 [cs].

\bibitem[Lester et~al.(2021)Lester, Al-Rfou, and Constant]{lester_power_2021}
Brian Lester, Rami Al-Rfou, and Noah Constant.
\newblock The {Power} of {Scale} for {Parameter}-{Efficient} {Prompt} {Tuning}, 2021.
\newblock arXiv:2104.08691 [cs].

\bibitem[Li and Liang(2021)]{li_prefix-tuning_2021}
Xiang~Lisa Li and Percy Liang.
\newblock Prefix-{Tuning}: {Optimizing} {Continuous} {Prompts} for {Generation}.
\newblock In \emph{Proceedings of the 59th {Annual} {Meeting} of the {Association} for {Computational} {Linguistics} and the 11th {International} {Joint} {Conference} on {Natural} {Language} {Processing} ({Volume} 1: {Long} {Papers})}, pages 4582--4597, Online, 2021. Association for Computational Linguistics.

\bibitem[Li and Xu(2023)]{li_trade-off_2023}
Yanxi Li and Chang Xu.
\newblock Trade-off between {Robustness} and {Accuracy} of {Vision} {Transformers}.
\newblock In \emph{2023 {IEEE}/{CVF} {Conference} on {Computer} {Vision} and {Pattern} {Recognition} ({CVPR})}, pages 7558--7568, Vancouver, BC, Canada, 2023. IEEE.

\bibitem[Lialin et~al.(2024)Lialin, Deshpande, Yao, and Rumshisky]{lialin_scaling_2024}
Vladislav Lialin, Vijeta Deshpande, Xiaowei Yao, and Anna Rumshisky.
\newblock Scaling {Down} to {Scale} {Up}: {A} {Guide} to {Parameter}-{Efficient} {Fine}-{Tuning}, 2024.
\newblock arXiv:2303.15647 [cs].

\bibitem[Little et~al.(2022)Little, Weylandt, and Allen]{little_fairness_2022}
Camille~Olivia Little, Michael Weylandt, and Genevera~I. Allen.
\newblock To the {Fairness} {Frontier} and {Beyond}: {Identifying}, {Quantifying}, and {Optimizing} the {Fairness}-{Accuracy} {Pareto} {Frontier}, 2022.
\newblock arXiv:2206.00074 [stat].

\bibitem[Liu et~al.(2022)Liu, Tam, Muqeeth, Mohta, Huang, Bansal, and Raffel]{liu_few-shot_2022}
Haokun Liu, Derek Tam, Mohammed Muqeeth, Jay Mohta, Tenghao Huang, Mohit Bansal, and Colin Raffel.
\newblock Few-{Shot} {Parameter}-{Efficient} {Fine}-{Tuning} is {Better} and {Cheaper} than {In}-{Context} {Learning}, 2022.
\newblock arXiv:2205.05638 [cs].

\bibitem[Liu et~al.(2023)Liu, Xu, Ji, and Chan]{liu_twins_2023}
Ziquan Liu, Yi Xu, Xiangyang Ji, and Antoni~B. Chan.
\newblock {TWINS}: {A} {Fine}-{Tuning} {Framework} for {Improved} {Transferability} of {Adversarial} {Robustness} and {Generalization}.
\newblock In \emph{2023 {IEEE}/{CVF} {Conference} on {Computer} {Vision} and {Pattern} {Recognition} ({CVPR})}, pages 16436--16446, Vancouver, BC, Canada, 2023. IEEE.

\bibitem[Madry et~al.(2019)Madry, Makelov, Schmidt, Tsipras, and Vladu]{madry_towards_2019}
Aleksander Madry, Aleksandar Makelov, Ludwig Schmidt, Dimitris Tsipras, and Adrian Vladu.
\newblock Towards {Deep} {Learning} {Models} {Resistant} to {Adversarial} {Attacks}, 2019.
\newblock arXiv:1706.06083 [stat].

\bibitem[Mahabadi et~al.(2021)Mahabadi, Henderson, and Ruder]{mahabadi_compacter_2021}
Rabeeh~Karimi Mahabadi, James Henderson, and Sebastian Ruder.
\newblock Compacter: {Efficient} {Low}-{Rank} {Hypercomplex} {Adapter} {Layers}, 2021.
\newblock arXiv:2106.04647 [cs].

\bibitem[Ng(2004)]{ng_feature_2004}
Andrew~Y. Ng.
\newblock Feature selection, \textit{{L}}$_{\textrm{1}}$ vs. \textit{{L}}$_{\textrm{2}}$ regularization, and rotational invariance.
\newblock In \emph{Twenty-first international conference on {Machine} learning - {ICML} '04}, page~78, Banff, Alberta, Canada, 2004. ACM Press.

\bibitem[Nguyen and Le(2024)]{nguyen_adapters_2024}
Tuc Nguyen and Thai Le.
\newblock Adapters {Mixup}: {Mixing} {Parameter}-{Efficient} {Adapters} to {Enhance} the {Adversarial} {Robustness} of {Fine}-tuned {Pre}-trained {Text} {Classifiers}, 2024.
\newblock arXiv:2401.10111 [cs].

\bibitem[Papernot et~al.(2015)Papernot, McDaniel, Jha, Fredrikson, Celik, and Swami]{papernot_limitations_2015}
Nicolas Papernot, Patrick McDaniel, Somesh Jha, Matt Fredrikson, Z.~Berkay Celik, and Ananthram Swami.
\newblock The {Limitations} of {Deep} {Learning} in {Adversarial} {Settings}, 2015.
\newblock arXiv:1511.07528 [cs].

\bibitem[Peng et~al.(2019)Peng, Bai, Xia, Huang, Saenko, and Wang]{peng_moment_2019}
Xingchao Peng, Qinxun Bai, Xide Xia, Zijun Huang, Kate Saenko, and Bo Wang.
\newblock Moment {Matching} for {Multi}-{Source} {Domain} {Adaptation}, 2019.
\newblock arXiv:1812.01754 [cs].

\bibitem[Recht et~al.(2018)Recht, Roelofs, Schmidt, and Shankar]{recht_cifar-10_2018}
Benjamin Recht, Rebecca Roelofs, Ludwig Schmidt, and Vaishaal Shankar.
\newblock Do {CIFAR}-10 {Classifiers} {Generalize} to {CIFAR}-10?, 2018.
\newblock arXiv:1806.00451 [cs].

\bibitem[Ridnik et~al.(2021)Ridnik, Ben-Baruch, Noy, and Zelnik-Manor]{ridnik_imagenet-21k_2021}
Tal Ridnik, Emanuel Ben-Baruch, Asaf Noy, and Lihi Zelnik-Manor.
\newblock {ImageNet}-{21K} {Pretraining} for the {Masses}, 2021.
\newblock arXiv:2104.10972 [cs].

\bibitem[Tay et~al.(2022)Tay, Dehghani, Rao, Fedus, Abnar, Chung, Narang, Yogatama, Vaswani, and Metzler]{tay_scale_2022}
Yi Tay, Mostafa Dehghani, Jinfeng Rao, William Fedus, Samira Abnar, Hyung~Won Chung, Sharan Narang, Dani Yogatama, Ashish Vaswani, and Donald Metzler.
\newblock Scale {Efficiently}: {Insights} from {Pre}-training and {Fine}-tuning {Transformers}, 2022.
\newblock arXiv:2109.10686 [cs].

\bibitem[Tsipras et~al.(2019)Tsipras, Santurkar, Engstrom, Turner, and Madry]{tsipras_robustness_2019}
Dimitris Tsipras, Shibani Santurkar, Logan Engstrom, Alexander Turner, and Aleksander Madry.
\newblock Robustness {May} {Be} at {Odds} with {Accuracy}, 2019.
\newblock arXiv:1805.12152 [stat].

\bibitem[Vaswani et~al.(2023)Vaswani, Shazeer, Parmar, Uszkoreit, Jones, Gomez, Kaiser, and Polosukhin]{vaswani_attention_2023}
Ashish Vaswani, Noam Shazeer, Niki Parmar, Jakob Uszkoreit, Llion Jones, Aidan~N. Gomez, Lukasz Kaiser, and Illia Polosukhin.
\newblock Attention {Is} {All} {You} {Need}, 2023.
\newblock arXiv:1706.03762 [cs].

\bibitem[Wah et~al.(2011)Wah, Branson, Welinder, Perona, and Belongie]{wah_caltech-ucsd_2011}
Catherine Wah, Steve Branson, Peter Welinder, Pietro Perona, and Serge Belongie.
\newblock The {Caltech}-{UCSD} {Birds}-200-2011 {Dataset}, 2011.

\bibitem[Xu et~al.(2023)Xu, Zhang, and Kankanhalli]{xu_autolora_2023}
Xilie Xu, Jingfeng Zhang, and Mohan Kankanhalli.
\newblock {AutoLoRa}: {A} {Parameter}-{Free} {Automated} {Robust} {Fine}-{Tuning} {Framework}, 2023.
\newblock arXiv:2310.01818 [cs].

\bibitem[Zaken et~al.(2022)Zaken, Ravfogel, and Goldberg]{zaken_bitfit_2022}
Elad~Ben Zaken, Shauli Ravfogel, and Yoav Goldberg.
\newblock {BitFit}: {Simple} {Parameter}-efficient {Fine}-tuning for {Transformer}-based {Masked} {Language}-models, 2022.
\newblock arXiv:2106.10199 [cs].

\bibitem[Zhai et~al.(2020)Zhai, Puigcerver, Kolesnikov, Ruyssen, Riquelme, Lucic, Djolonga, Pinto, Neumann, Dosovitskiy, Beyer, Bachem, Tschannen, Michalski, Bousquet, Gelly, and Houlsby]{zhai_large-scale_2020}
Xiaohua Zhai, Joan Puigcerver, Alexander Kolesnikov, Pierre Ruyssen, Carlos Riquelme, Mario Lucic, Josip Djolonga, Andre~Susano Pinto, Maxim Neumann, Alexey Dosovitskiy, Lucas Beyer, Olivier Bachem, Michael Tschannen, Marcin Michalski, Olivier Bousquet, Sylvain Gelly, and Neil Houlsby.
\newblock A {Large}-scale {Study} of {Representation} {Learning} with the {Visual} {Task} {Adaptation} {Benchmark}, 2020.
\newblock arXiv:1910.04867 [cs].

\bibitem[Zhang et~al.(2019)Zhang, Yu, Jiao, Xing, Ghaoui, and Jordan]{zhang_theoretically_2019}
Hongyang Zhang, Yaodong Yu, Jiantao Jiao, Eric Xing, Laurent~El Ghaoui, and Michael Jordan.
\newblock Theoretically {Principled} {Trade}-off between {Robustness} and {Accuracy}.
\newblock In \emph{Proceedings of the 36th {International} {Conference} on {Machine} {Learning}}, pages 7472--7482. PMLR, 2019.
\newblock ISSN: 2640-3498.

\bibitem[Zhou et~al.(2018)Zhou, Lapedriza, Khosla, Oliva, and Torralba]{zhou_places_2018}
Bolei Zhou, Agata Lapedriza, Aditya Khosla, Aude Oliva, and Antonio Torralba.
\newblock Places: {A} 10 {Million} {Image} {Database} for {Scene} {Recognition}.
\newblock \emph{IEEE Transactions on Pattern Analysis and Machine Intelligence}, 40\penalty0 (6):\penalty0 1452--1464, 2018.

\end{thebibliography}
}

\appendix
\newpage  
\section{Supplementary Material}\label{sec:appendix}

\subsection{Datasets}\label{sec:app:datasets}
\noindent \textbf{\texttt{CIFAR10}}~\cite{recht_cifar-10_2018} is widely used for image classification. It contains $10$ classes with $60,000$ images ($50,000$ for training and $10,000$ for testing). Its small resolution ($32\times32$ pixels) and balanced class distribution make it a common benchmark for evaluating adversarial robustness. \texttt{CIFAR10} has significantly fewer classes and a lower level of visual complexity compared to \texttt{ImageNet21-k}. This enables us to study how fine-tuning on simpler datasets change model robustness inherited from pre-training. 

\noindent \textbf{\texttt{CIFAR100.}} \texttt{CIFAR100}~\cite{krizhevsky_cifar-10_2009} extends \texttt{CIFAR10} to $100$ classes, each containing $600$ images ($500$ for training, $100$ for testing). While it shares the same low-resolution format, \texttt{CIFAR100} introduces a more fine-grained classification task. The increased class diversity and hierarchical structure (coarse and fine labels) make it a more complex dataset but still much smaller in scale compared to \texttt{ImageNet-21k}. 

\noindent \textbf{\texttt{Caltech256.}} \texttt{Caltech256}~\cite{griffin_caltech_2022} comprises $256$ classes with $30,607$ images, offering significantly more class diversity than CIFAR datasets. It has a minimum of 80 images per class. \texttt{Caltech256} contains higher-resolution images with more natural object variations, making it more similar to \texttt{ImageNet-21k} in terms of complexity and scale. With this, we can better understand how fine-tuning on a moderately large dataset with varied classes affects robustness. 

\noindent \textbf{\texttt{CUB-200-2011.}} \texttt{CUB200}~\cite{wah_caltech-ucsd_2011} is a fine-grained classification dataset containing $11,788$ images across $200$ bird species. Unlike broader classification datasets, \texttt{CUB200} focuses on a single semantic category (birds), making it an important benchmark for studying adversarial robustness in tasks where pre-trained models are fine-tuned on more specialized, domain-specific knowledge. Since \texttt{ImageNet-21k} includes bird species in its taxonomy, this dataset allows us to explore how fine-tuning on a subdomain of the pre-training distribution impacts robustness.

\noindent \textbf{\texttt{Stanford Dogs.}} \texttt{Stanforddogs}~\cite{khosla_novel_2011} is another fine-grained classification dataset with $22,000$ images of $120$ dog breeds. Similar to CUB, it provides a challenging adversarial benchmark due to the subtle intra-class variations among breeds. Since \texttt{ImageNet-21k} also contains dog breeds, this dataset enables us to investigate whether fine-tuning on a narrower but related distribution affects the robustness inherited from pre-training. 

\noindent \textbf{\texttt{DomainNet}.} \texttt{DomainNet}~\cite{peng_moment_2019} is a large-scale domain adaptation dataset of $586,575$ images, containing six different domains: clip art, info graph, painting, quick draw, real, and sketch. ImageNet-21k primarily contains real-world images, making \texttt{DomainNet} an effective benchmark to test how well fine-tuned models generalize when faced with significant distributional changes, particularly when trained on one domain and tested on others. 

\subsection{Trade-off Space of Fine-Tuning}
\begin{figure}[ht]
  \centering
  \begin{subfigure}{0.49\columnwidth}
    \includegraphics[width=\columnwidth]{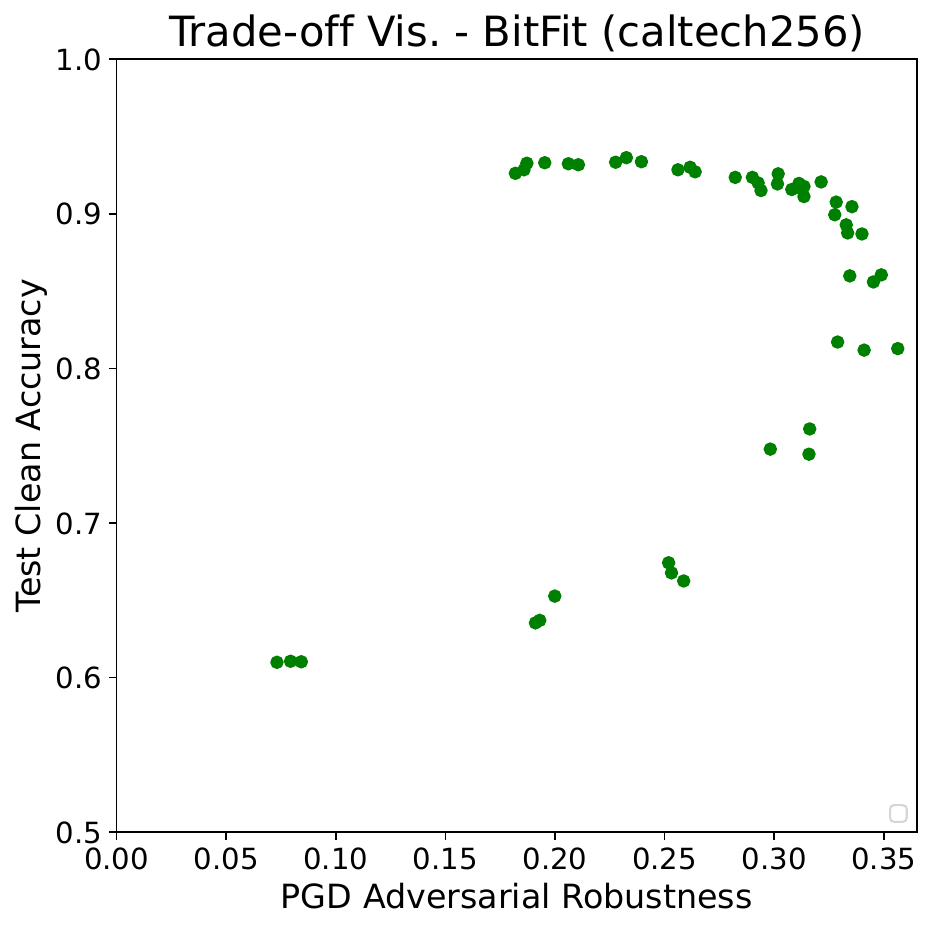}
    \caption{BitFit (Caltech256)}
    \label{fig:pareto_bitfit_caltech256}
  \end{subfigure}
  \hfill
  \begin{subfigure}{0.49\columnwidth}
    \includegraphics[width=\columnwidth]{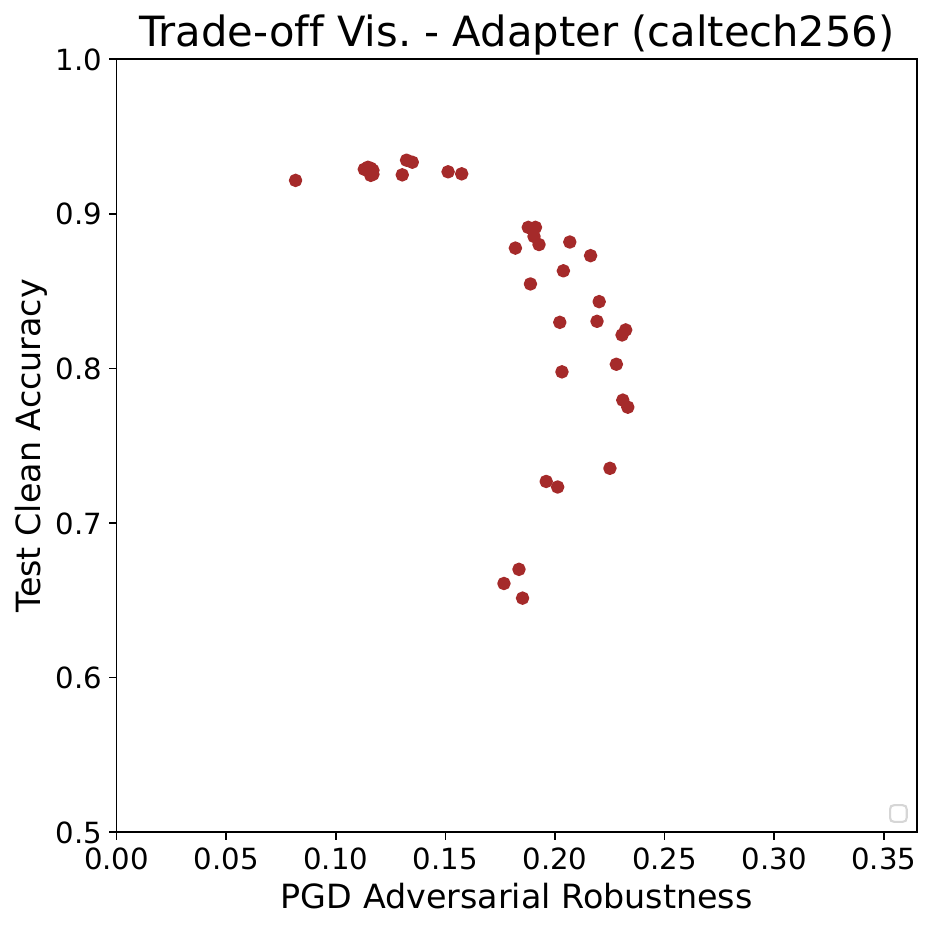}
    \caption{Adapter (Caltech256)}
    \label{fig:pareto_adapter_caltech256}
  \end{subfigure}

  \begin{subfigure}{0.49\columnwidth}
    \includegraphics[width=\columnwidth]{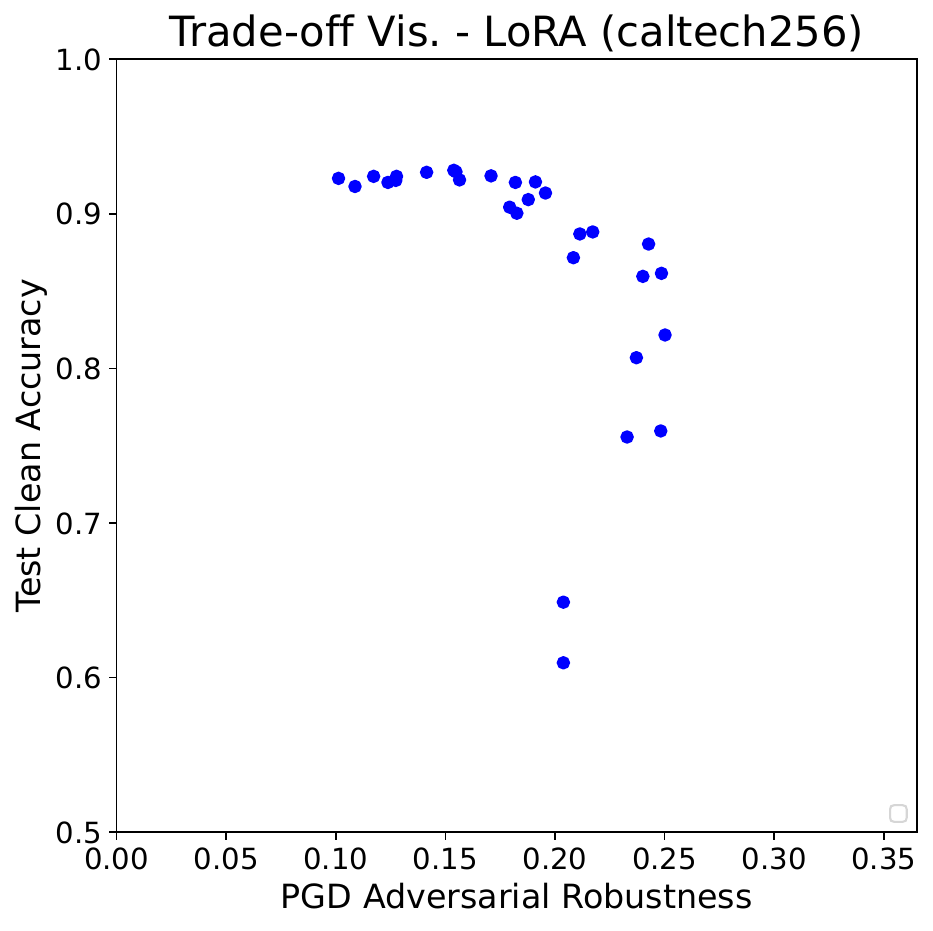}
    \caption{LoRA (Caltech256)}
    \label{fig:pareto_lora_caltech256}
  \end{subfigure}
  \hfill
  \begin{subfigure}{0.49\columnwidth}
    \includegraphics[width=\columnwidth]{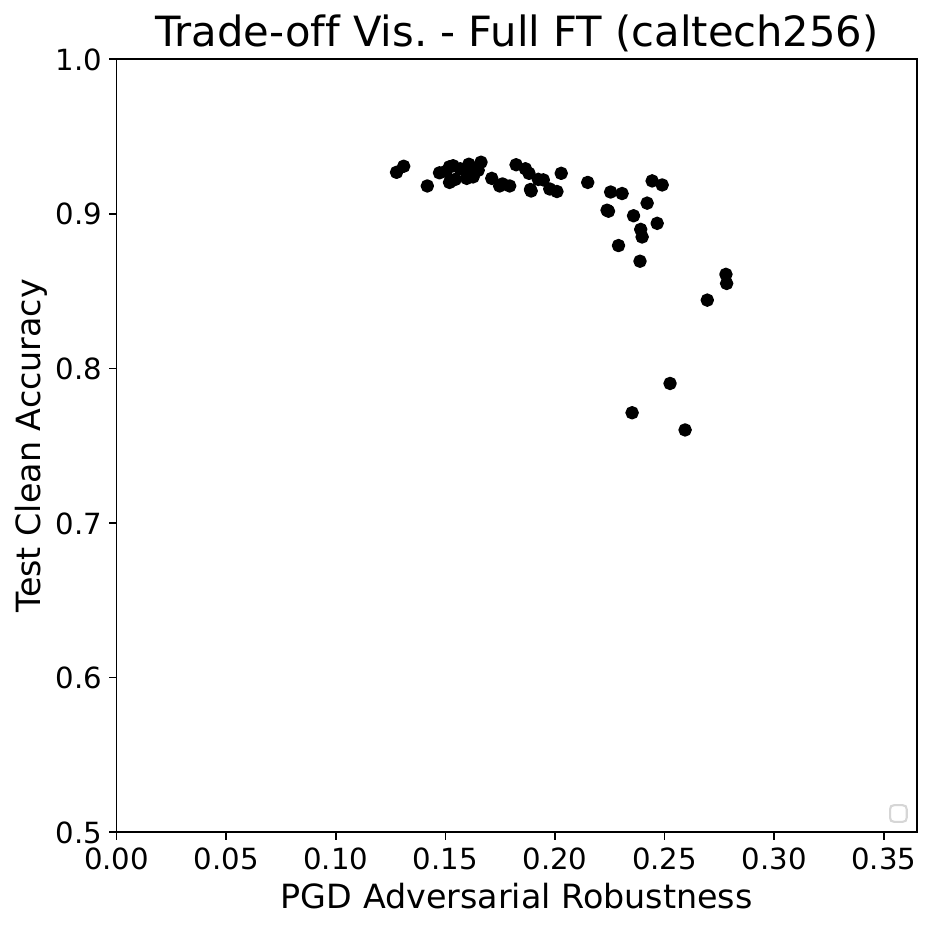}
    \caption{Full fine-tuning (Caltech256)}
    \label{fig:pareto_full_fine-tune_caltech256}
  \end{subfigure}
  
  \caption{Trade-off visualization for Caltech256. The dots are corresponding to different time stamps during training (from bottom left to upper right to upper left as time goes on).}
    \label{fig:pareto_curves}
\end{figure}

We present the trade-off space between adversarial robustness and accuracy in \autoref{fig:pareto_curves}. This corresponds to the training curves shown in \autoref{fig:training_curves}. 

\subsection{Ablation Study} \label{sec:app:ablation}
We perform an ablation study on the number and location of trainable parameters and learning rate. We conduct experiment sweeps with final model checkpoints on 1) LoRA rank (r $\in$ [1,20]), 2) Adapter reduction factor ($d \in$ [4,32]), and 3) update location (attention vs. FNN). The results show that robustness does not consistently correlate with the number of trainable parameters or layers updated. This suggests that these factors alone do not explain robustness differences. However, the results all show low adversarial accuracy with small variances. It led us to our design choise: tracking changes over training steps offer deeper insights. 

\begin{figure}[t]
  \centering
  \includegraphics[width=0.8\linewidth]{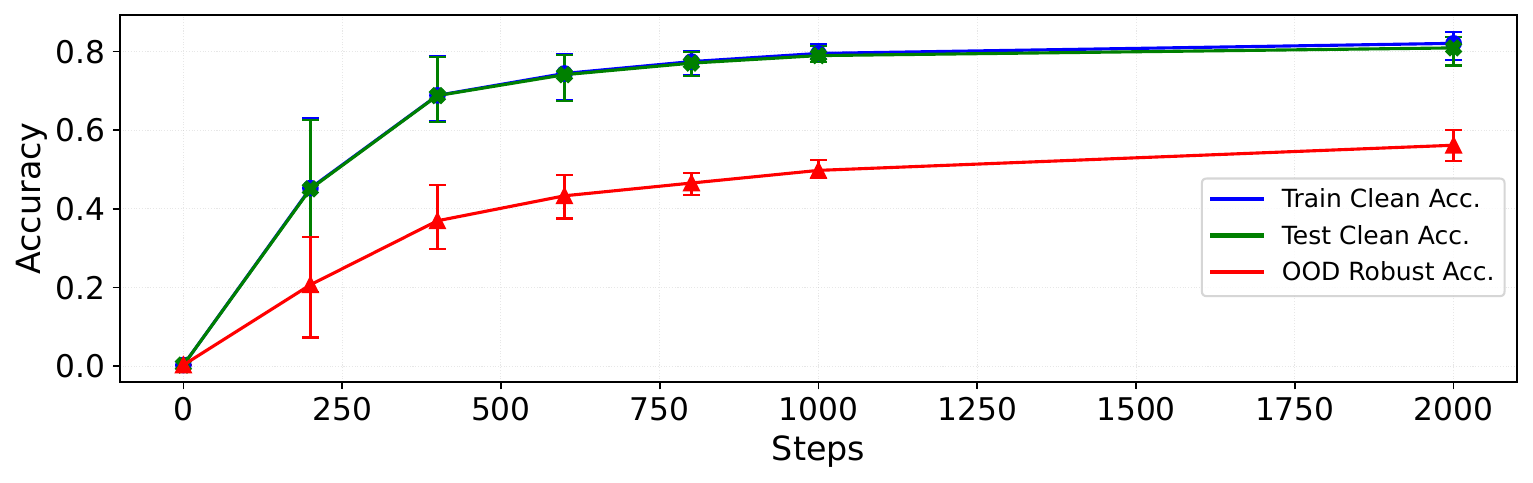}
   \caption{The OOD robustness of models fine-tuned by LoRA with different learning rates on the \textit{``real"} domain and evaluated on 5 other domains.}
   \label{fig:domainnet_lr_ablation}
\end{figure}

In addition, we also study how different learning rates affect OOD robustness during training. We track OOD robustness while fine-tuning models with varying learning rates---$\{1e-4, 5e-4, 1e-3, 5e-3\}$. As shown in \autoref{fig:domainnet_lr_ablation}, learning rate does impact accuracy and robustness, especially during the early steps, but show consistent trend as claimed across steps and converge in the end. 

\subsection{Hyperparameters}\label{sec:hyperparameters}
Grid search is used to find optimal training hyperparameters: base learning rate in $\{1e-4, 1e-5, 3e-5, 5e-5\}$, base weight decay in $\{1-e2, 1-e3\}$, and the adjustment ratio for each fine-tuning strategy in $\{1, 10, 5, 10, 2, 2, 3\}$ (corresponding to the order of the strategies shown in \autoref{tab:hyperparameters_std_datasets}). These choices are based on previous literature~\cite{zaken_bitfit_2022,hu_lora_2021,houlsby_parameter-efficient_2019} for different fine-tuning methods and downstream datasets. They all have comparable scale of trainable parameters (in percentage), except for full fine-tuning:$\{100, 0.01, 1.19, 0.13, 2.03, 0.07, 0.07 \}$. Due to the large size of DomainNet~\cite{peng_moment_2019}, we consistently set the base weight decay to be $1e-2$. The specific optimal training hyperparameters can be found in \autoref{tab:hyperparameters_std_datasets} and \autoref{tab:hyperparameters_ood}.  

\begin{table*}[htbp]
\centering
\small
\begin{tabular}{l|c|c|c|c|c}
\hline
\multirow{2}{*}{\makecell[l]{Fine-Tuning \\ Methods}} & \multicolumn{5}{c}{Learning Rate / Weight Decay for Adv Exps.} \\
\cline{2-6}
 & CIFAR10 & CIFAR100 & CUB200 & Caltech256 & Stanford Dogs \\
\hline
Full Fine-tune & 3e-5/1e-3 & 5e-5/1e-3 & 5e-5/1e-3 & 1e-4/1e-3 & 1e-4/1e-3 \\
Linear Probe & 1e-5/1e-3 & 1e-5/1e-2 & 5e-6/1e-3 & 1e-5/1e-3 & 1e-5/1e-3 \\
LoRA & 5e-4/1e-2 & 5e-4/1e-2 & 2.5e-4/1e-2 & 5e-4/1e-3 & 5e-5/1e-2 \\
BitFit & 1e-5/1e-4 & 1e-5/1e-3 & 5e-6/1e-4 & 1e-5/1e-3 & 1e-5/1e-3 \\
Adapter & 1e-4/1e-3 & 2e-4/1e-2 & 2e-5/1e-2 & 2e-4/1e-2 & 2e-4/1e-3 \\
Compacter & 2e-4/1e-3 & 2e-4/1e-3 & 1e-4/1e-3 & 2e-4/1e-3 & 2e-4/1e-3 \\
(IA)$^3$ & 1.5e-4/1e-3 & 3e-4/1e-2 & 3e-4/1e-3 & 3e-4/1e-3 & 1.5e-4/1e-3 \\
\hline
\end{tabular}
\caption{Strategy configurations with datasets (Adv)}
\label{tab:hyperparameters_std_datasets}
\end{table*}

\begin{table*}[htbp]
\centering
\small
\begin{tabular}{l|c|c|c|c|c|c}
\hline
\multirow{2}{*}{\makecell[l]{Fine-Tuning \\ Methods}} & \multicolumn{6}{c}{Learning Rate for OOD Exps.} \\
\cline{2-7}
  & Clipart & Infograph & Painting & Quickdraw & Real & Sketch \\
\hline
Full Fine-tune & 1e-4 & 1e-4 & 1e-4 & 1e-4 & 1e-4 & 1e-4 \\
Linear Probe & 1e-3 & 1e-3 & 1e-3 & 1e-3 & 5e-4 & 1e-3 \\
LoRA & 5e-4 & 2.5e-4 & 5e-4 & 2.5e-4 & 5e-4 & 5e-4 \\
BitFit & 1e-3 & 1e-3 & 5e-4 & 1e-3 & 5e-4 & 1e-3 \\
Adapter & 2e-4 & 2e-4 & 2e-4 & 1e-4 & 2e-4 & 2e-4 \\
Compacter & 2e-4 & 2e-4 & 2e-4 & 2e-4 & 2e-4 & 2e-4 \\
(IA)$^3$ & 3e-4 & 3e-4 & 3e-4 & 3e-4 & 3e-4 & 3e-4 \\
\hline
\end{tabular}
\caption{Strategy configurations with datasets (OOD)}
\label{tab:hyperparameters_ood}
\end{table*}

\subsection{Dataset Scale}\label{sec:app:data_scale}
\begin{figure}[t]
    \centering
\includegraphics[width=1\columnwidth]{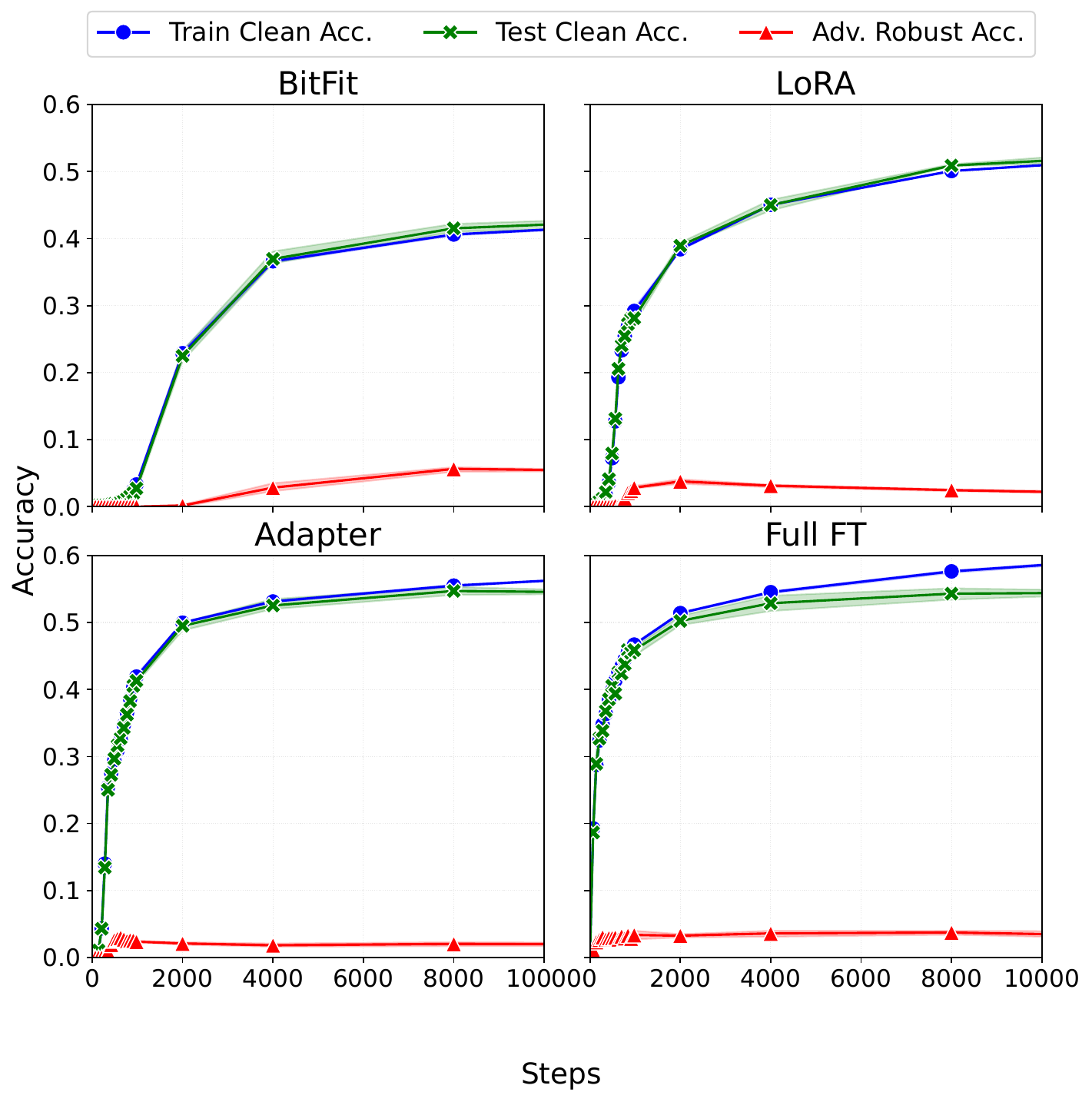}
    \caption{Continuous evaluation of training accuracy (blue), test accuracy (green), and adversarial robustness (red) across backpropagation steps (truncated at 10,000 steps) on Places365~\cite{zhou_places_2018}.}
    \label{fig:places365_training_curves}
\end{figure}
PEFT is especially relevant in low-data regimes. Our main experiments focus on datasets with 10k to 60k samples, covering diverse degrees of task complexity in terms of number of classes, class separation, and similarity to upstream data. To complement this analysis and assess the effect of data scale, we also include experiments on Places365~\cite{zhou_places_2018}, a medium-scale dataset with approximately 1.8 million samples. As shown in \autoref{fig:places365_training_curves}, models achieve relatively low clean accuracy (preaking around 50\%) and limited adversarial robustness (less than $5\%$). Due to this overall weak performance, the trend of adversarial robustness and the differences of robustness across fine-tuning strategies are difficult to distinguish.

\subsection{Decomposition of Fine-Tuning}\label{sec:app:decomp}
As described in \autoref{method:decomp_peft}, we focus on decomposing PEFT methods along two directions: information location and mechanisms, each having four components. The decomposition for five PEFT methods can be found in \autoref{tab:peft_space}.
\begin{table*}[ht]
\centering
    \begin{tabular}{l c c c c  c c c c}
    \Xhline{3\arrayrulewidth}
     & \multicolumn{4}{c}{\textbf{Information Location}}  & \multicolumn{4}{c}{\textbf{Mechanism}} \\
    \cline{2-5}
    \cline{6-9}
    \makecell{PEFT \\ Strategies} & Attn & FFN & Rep. & Bias & \makecell{Proj. \\ Layers} & \makecell{Matrix \\ Reparam} &  \makecell{Element-wise \\ Mult.} & \makecell{Direct \\ Update} \\
    \hline
    LoRA & $\bullet$ & $\circ$ & $\circ$ & $\circ$ & $\circ$ & $\bullet$ & $\circ$ & $\circ$ \\
    IA3 & $\circ$ & $\circ$ & $\bullet$ & $\circ$ & $\circ$ & $\circ$ & $\bullet$ & $\circ$ \\
    Adapter & $\circ$ & $\circ$ & $\bullet$ & $\circ$ & $\bullet$ & $\circ$ & $\circ$ & $\circ$ \\
    Compacter & $\circ$ & $\circ$ & $\bullet$ & $\circ$ & $\bullet$ & $\bullet$ & $\circ$ & $\circ$ \\
    BitFit & $\bullet$ & $\bullet$ & $\circ$ & $\bullet$ & $\circ$ & $\circ$ & $\circ$ & $\bullet$ \\
    \Xhline{3\arrayrulewidth}
    \end{tabular}

\caption{The space of PEFT strategies in terms of information location and underlying mechanisms}
\label{tab:peft_space}
\end{table*}

\end{document}